\documentclass[10pt]{article} 
\usepackage{times}
\usepackage[left=1.5in, right=1.5in, top=1in, bottom=1in]{geometry}
\usepackage{breakcites}
\usepackage{graphicx}
\usepackage{textcomp}
\usepackage{xcolor}
\def\BibTeX{{\rm B\kern-.05em{\sc i\kern-.025em b}\kern-.08em
    T\kern-.1667em\lower.7ex\hbox{E}\kern-.125emX}}

\usepackage{amsmath}
\usepackage[]{algorithm2e}
\usepackage{algpseudocode}
\SetKwRepeat{Do}{do}{while}
\usepackage{graphicx}  
\usepackage{float}
\usepackage{bm}      
\usepackage{epstopdf}
\usepackage{subcaption}
\usepackage{amssymb}   
\usepackage{amsmath}
\usepackage{booktabs}       
\usepackage{hyperref}
\usepackage{parskip}
\usepackage{todonotes}
\usepackage{xcolor}
\usepackage{placeins}

\begin{document}

\title{Latent Space Exploration Using Generative Kernel PCA}

\author{David Winant, Joachim Schreurs and  Johan A. K. Suykens\\
Department of Electrical Engineering, ESAT-STADIUS,\\
KU Leuven. Kasteelpark Arenberg 10, B-3001 Leuven, Belgium\\
\texttt{\{david.winant,joachim.schreurs,johan.suykens\}@kuleuven.be} \\
}

\maketitle

\begin{abstract}
Kernel PCA is a powerful feature extractor which recently has seen a reformulation in the context of Restricted Kernel Machines (RKMs). These RKMs allow for a representation of kernel PCA in terms of hidden and visible units similar to Restricted Boltzmann Machines. This connection has led to insights on how to use kernel PCA in a generative procedure, called generative kernel PCA. In this paper, the use of generative kernel PCA for exploring latent spaces of datasets is investigated.  New points can be generated by gradually moving in the latent space, which allows for an interpretation of the components. Firstly, examples of this feature space exploration on three datasets are shown with one of them leading to an interpretable representation of ECG signals. Afterwards, the use of the tool in combination with novelty detection is shown, where the latent space around novel patterns in the data is explored. This helps in the interpretation of why certain points are considered as novel.
\end{abstract}
\section{Introduction}

Latent spaces provide a representation of data by embedding the data into an underlying vector space. Exploring these spaces allows for deeper insights in the structure of the data distribution, as well as understanding relationships between data points. Latent spaces are used for various purposes like latent space cartography~\cite{liu2019latent}, object shape generation~\cite{wu2016learning} or style-based generation~\cite{karras2019style}.  In this paper, the focus will be on how the synthesis of new data with generative methods can help with understanding the latent features extracted from a dataset.
In recent years, generative methods have become a hot research topic within the field of machine learning. Two of the most well-known examples include variational autoencoders (VAEs)~\cite{kingma2013auto} and Generative Adversarial Networks (GANs)~\cite{goodfellow2014generative}. An example of a real-world application of latent spaces using VAEs is shown in \cite{way2017extracting}, where deep convolututional VAEs are used to extract a biologically meaningful latent space from a cancer transcriptomes dataset. This latent space is used to explore hypothetical gene expression profiles of tumors and their reaction to possible treatments. Similarly disentangled variational autoencoders have been used to find an interpretable and explainable representation of ECG signals \cite{van2019interpretable}. Latent space exploration is also used for interpreting GANs, where interpolation between different images allows for the interpretation of the different features captured by the latent space, such as windows and curtains when working with datasets of bedroom images \cite{radford2015unsupervised}. Latent space models are especially appealing for the synthesis of plausible pseudo-data with certain desirable properties. If the latent space is disentangled or uncorrelated, it is easier to interpret the meaning of different components in the latent space. Therefore it is easier to generate examples with desired properties, e.g. we want to generate a new face with certain characteristics. More recently, the concept of latent space exploration with GANs has been further developed by introducing new couplings of the latent space to the architecture of the generative network, this allows for control of local features for image synthesis at different scales in a style-based design \cite{karras2019style}. These adaptations of GANs are known as Style-GANs. When applied to a facial dataset, the features can range from general face shape and hair style up to eyes, hair colour and mouth shape.

In this paper, kernel PCA is used as a generative mechanism~\cite{schreurs2018generative}. Kernel PCA, as first described in \cite{scholkopf1997kernel}, is a well-known feature extractor method often used for denoising and dimensionality reduction of datasets. Through the use of a kernel function it is a nonlinear extension to regular PCA by introducing an implicit, high dimensional latent feature space wherein the principal components are extracted. In \cite{suykens2017deep}, kernel PCA was cast within the framework of Restricted Kernel Machines (RKMs) which allows for an interpretation in terms of hidden and visible units similar to a type of generative neural network known as Restricted Boltzmann Machines (RBMs)~\cite{hinton2012practical}. This connection between kernel PCA and RBMs was later used to explore a generative mechanism for the kernel PCA~\cite{schreurs2018generative}. A tensor-based multi-view classification model was introduced in \cite{houthuys_tensor-based_nodate}. In \cite{GENRKM}, a multi-view generative model called Generative RKM (Gen-RKM) is proposed which uses explicit feature-maps in a novel training procedure for joint feature-selection and subspace learning.

The goal of this paper is to explore the latent feature space extracted by kernel PCA using a generative mechanism, in an effort to interpret the components. This has led to the development of a Matlab tool which can be used to visualise the latent space of the kernel PCA method along its principal components. The use of the tool is demonstrated on three different datasets: the MNIST digits dataset, the Yale Face database and the MIT-BIH Arrhythmia database. As a final illustration, feature space exploration is used in the context of novelty detection~\cite{hofmann2008kernel}, where the latent space around novel patterns in the data is explored. This to help the interpretation of why certain points are considered as novel.

In Section \ref{sec:Generative kernel PCA}, a brief review on generative kernel PCA is given. In Section \ref{sec:Experiments},  latent feature space exploration is demonstrated. Subsequently we will illustrate how latent feature space exploration can help in interpreting novelty detection in Section \ref{sec:Outlier visualisation}. The paper is concluded in Section \ref{sec:Conclusion}.

\section{Kernel PCA in the RKM framework}
\label{sec:Generative kernel PCA}

In this section, a short review on how kernel PCA can be used to generate new data is given, as introduced in~\cite{schreurs2018generative}. We start with the calculation of the kernel principal components for a $d$-dimensional dataset $\{ x_i\}^N_{i=1}$ with $N$ data points and for each data point $x_i \in \mathbb{R}^d$. Compared to regular PCA, kernel PCA first maps the input data to a high dimensional feature space $\mathcal{F}$ using a feature map $\phi(\cdot)$. In this feature space, regular PCA is performed on the points $\phi(x_i)$ for $ i = 1,\ldots,N$. By using a kernel function $k(x,y)$ defined as the inner product $( \phi(x)\cdot\phi(y))$, an explicit expression for $\phi(\cdot)$ can be avoided. Typical examples of such kernels are given by the Gaussian RBF kernel $k(x,y) = e^{-\|x-y\|_2^2/(2\sigma^2)}$ or the Laplace kernel $k(x,y) =e^{-\|x-y\|_2/\sigma}$, where $\sigma$ denotes the bandwidth. Finding the principal components amounts to solving the eigenvalue problem for the kernel matrix\footnote{For simplicity, the mapped data are assumed to be centered in $\mathcal{F}$. Otherwise, we have to go through the same algebra using $\tilde{\phi}(x) :=\phi(x)- \sum_{i=1}^N \phi(x_i)$. This is the same assumption as in~\cite{scholkopf1997kernel}.} $K$, with matrix elements $K_{i j}= ( \phi\left(\boldsymbol{x}_{i}\right) \cdot \phi\left(\boldsymbol{x}_{j}\right) )$. 
The eigenvalue problem for kernel PCA is stated as follows:

\begin{equation}\label{eq:kpca}
    K H^{\top} = H^{\top}\Lambda,
\end{equation}

where $H = [h_1,\dots,h_N] \in \mathbb{R}^{d \times N}$, the first $d \leq N$ components are used, is the matrix with the eigenvectors in each column and $\Lambda = \mathrm{diag}\{\lambda_1,\ldots,\lambda_d\}$ the matrix with the corresponding eigenvalues on the diagonal. In the framework of RKMs, the points $\phi(x_i)$ correspond to visible units $v_i$ and $h_i$ are the hidden units. As in \cite{schreurs2018generative}, the generative equation is given by:

\begin{equation}
\label{eq:vstar}
    v^{\star}= \phi(x^{\star}) = \left(\sum^N_{i=1}\phi(x_i)h^{\top}_i\right)h^{\star},
\end{equation}

where $h^{\star}$ represents a newly generated hidden unit and $v^{\star}$ the corresponding visible unit. Finding $x^{\star}$ in Eq. \eqref{eq:vstar} corresponds to the pre-image problem~\cite{honeine2011preimage}. In \cite{schreurs2018generative}, the authors give a possible solution by multiplying both sides with  $\phi(x_k)$, which gives the output of the kernel function for the generated point in the input space $x^{\star}$ and the data point $x_k$:

\begin{equation}
\label{eq:kHat}
    \hat{k}(x_k,x^{\star}) = \sum^N_{i=1}k(x_k,x_i)h_i^{\top}h^{\star}.
\end{equation}
The above equation can be seen as the similarity between the newly generated point $x^{\star}$ and $x_k$. This expression can be used in a kernel smoother approach to find an estimate $\hat{x}$ for the generated data point $x^{\star}$:
\begin{equation}
\label{eq:kernelSmooth}
    \hat{x} = \frac{\sum^S_{i=1}\tilde{k}(x_i,x^{\star})x_i}{\sum^S_{i=1}\tilde{k}(x_i,x^{\star})},    
\end{equation}
where $\tilde{k}(x_i,x^{\star})$ is the scaled similarity between 0 and 1 calculated in \eqref{eq:kHat} and $S$ the number of closest points based on the similarity $\tilde{k}\left(x_{i}, x^{\star}\right)$. Given a point in the latent space $h^{\star}$, we get an approximation for the corresponding point $\hat{x}$ in input space. This mechanism makes it possible to continuously explore the latent space.

\section{Experiments}
\label{sec:Experiments}

Our goal is to use generative kernel PCA to explore the latent space. Therefore a tool\footnote{Matlab code for the latent space exploration tool is available at \url{https://www.esat.kuleuven.be/stadius/E/software.php}. } is developed where generative kernel PCA can easily be utilised for new datasets. First  kernel PCA is performed to find the hidden features of the dataset. After choosing an initial hidden unit as starting point, the values are varied for each component of the hidden unit to explore the latent space. The corresponding newly generated data point in the input space is estimated using the kernel smoother approach.

\begin{figure}[h]
    \centering
    \includegraphics[width=1.0\linewidth]{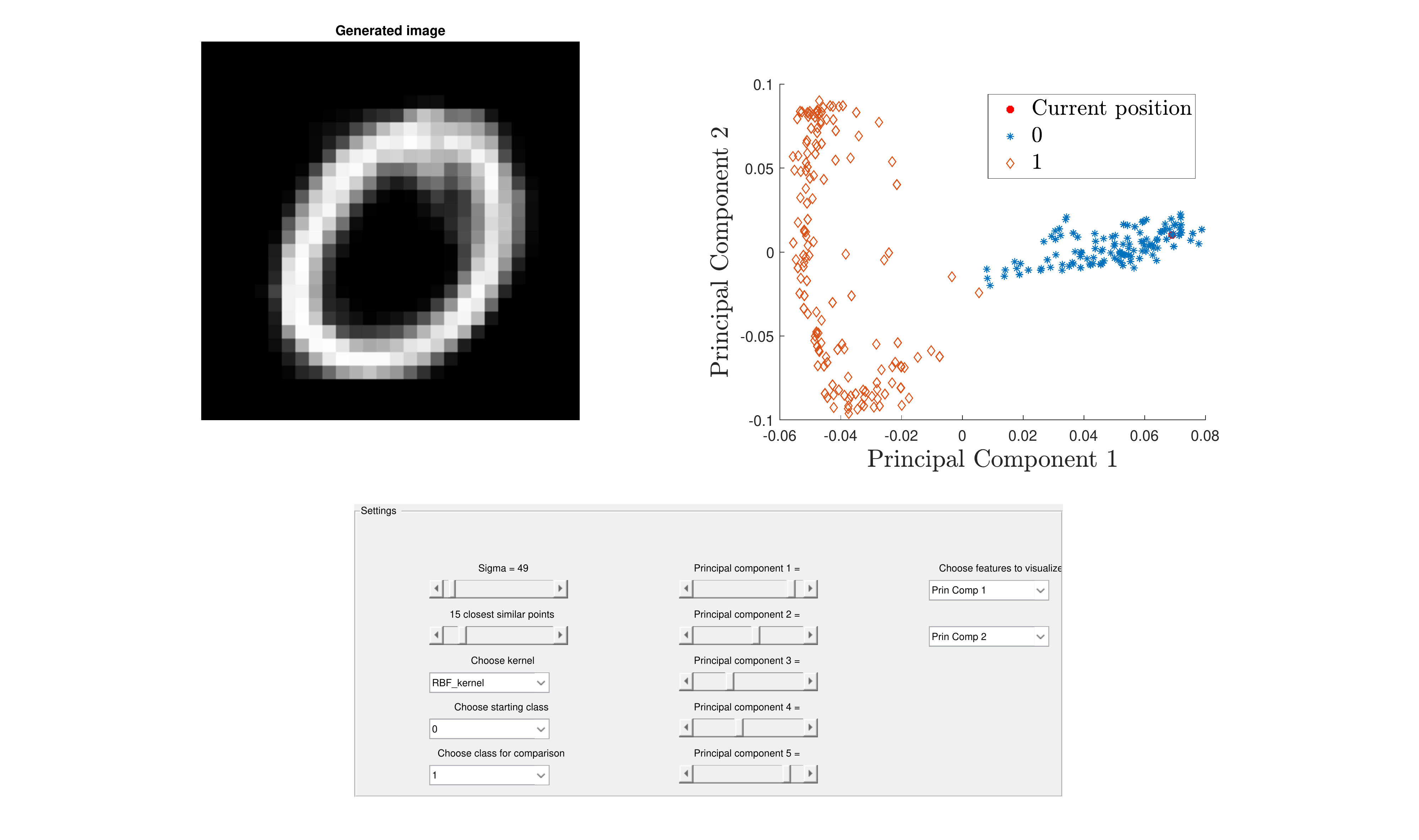}
    \caption{Interface of the Matlab tool for exploring the latent space. At the bottom, the parameter values and position in the latent space can be chosen. In the top right the latent space along two selected principal components is shown and on the left the newly generated data point in the input space is visualised.}
    \label{fig:GUI}
\end{figure}

In the tool, a partial visualisation of the latent space projected onto two principal components is shown. We continuously vary the values of the components of the selected hidden unit. This allows the exploration of the extracted latent space by visualising the resulting variation in the input space. The ability to perform incremental variations aids interpretation of the meaning encoded in the latent space along a chosen direction. In Fig. \ref{fig:GUI}, the interface of our tool is shown .

\newpage

\subsubsection{MNIST Handwritten Digits}\hfill

\begin{figure}[h]
\centering
\begin{subfigure}{0.49\textwidth}
    \centering
    \includegraphics[width=1.0\linewidth]{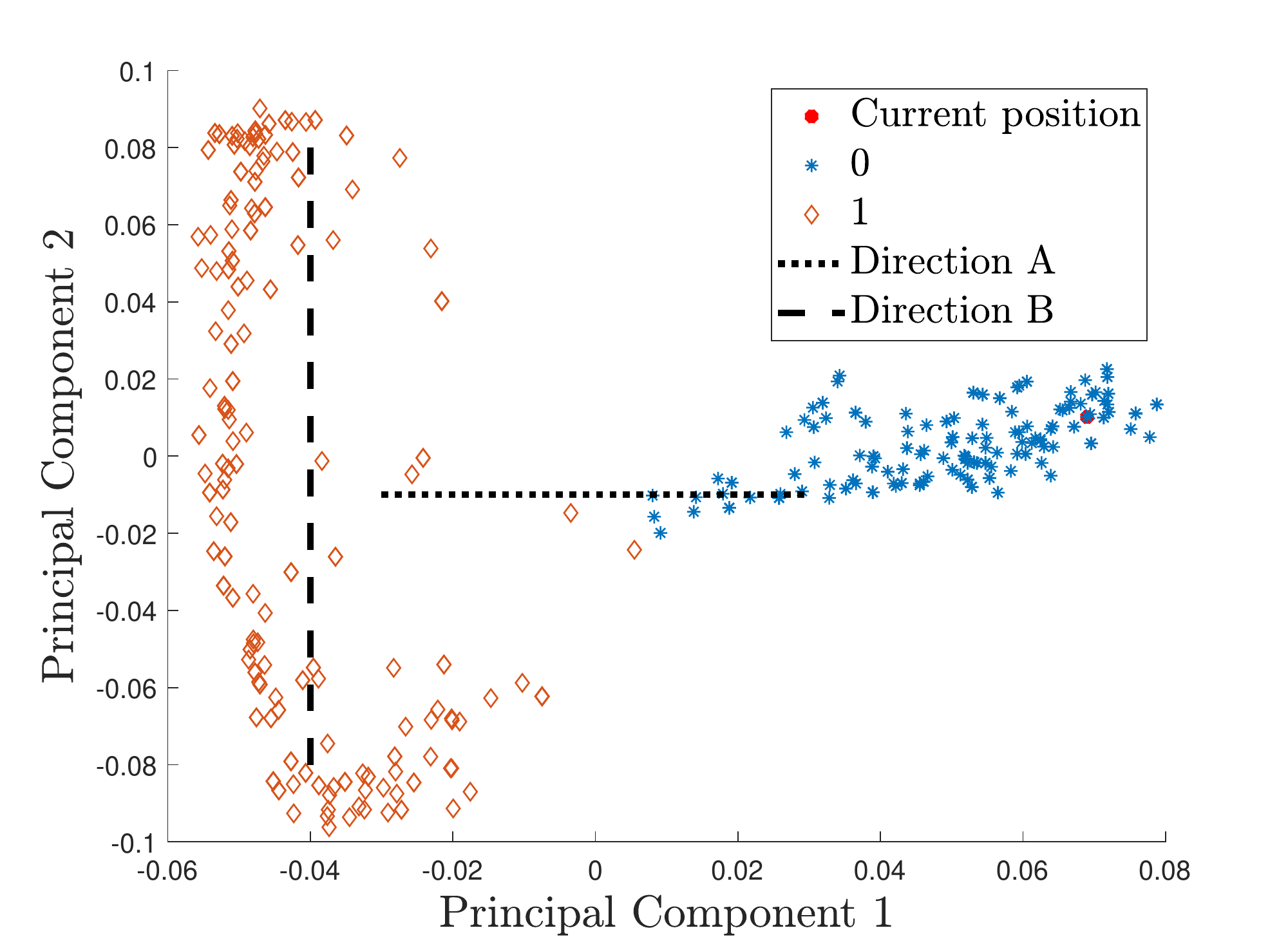}
    \caption{}
    \label{fig:MNIST_feat_space_12}
\end{subfigure}
\begin{subfigure}{0.49\textwidth}
\centering
    \includegraphics[width=1.0\linewidth]{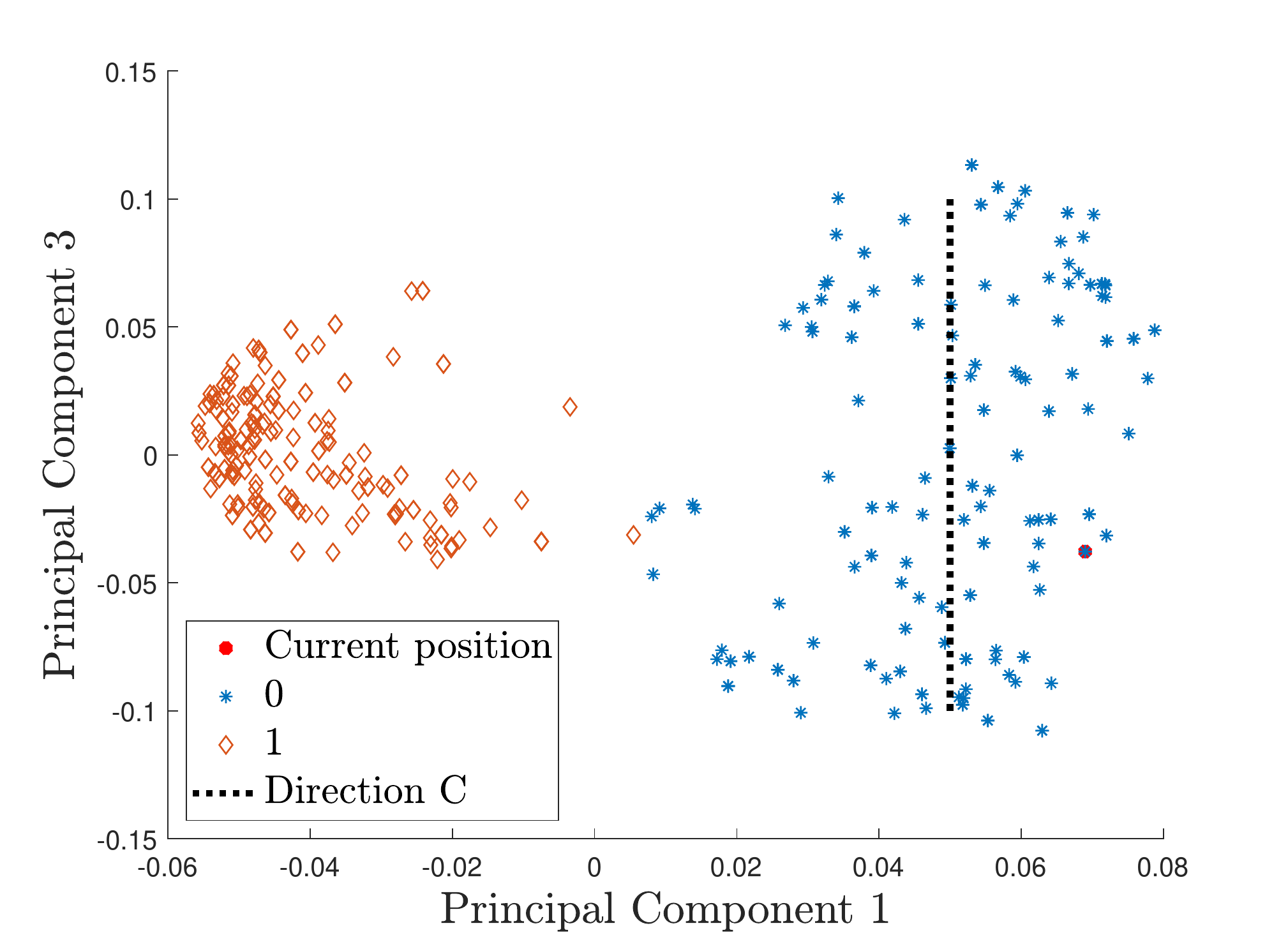}
    \caption{}
    \label{fig:MNIST_feat_space_13}
\end{subfigure}
\caption{Latent space of the MNIST digits dataset for the digits 0 and 1. The dotted lines indicate the direction along which new data points are generated. (a) Data projected on the first two principal components (b) Data projected on the first and third principal component. }
\label{fig:MNIST_exploration}
\end{figure}

As an example, the latent space of the MNIST handwritten digits dataset~\cite{lecun1998gradient} is explored, where 1000 data points each of digits zero and one are sampled. A Gaussian kernel with bandwidth $\sigma^2=50$, $S = 15$  and number of components $d=10$. In Fig. \ref{fig:MNIST_exploration}, the latent space is shown along the first two principal components as well as the first and third components.

\begin{figure}
\centering
\begin{subfigure}[t]{.12\textwidth}
  \centering
  \includegraphics[width=.8\linewidth]{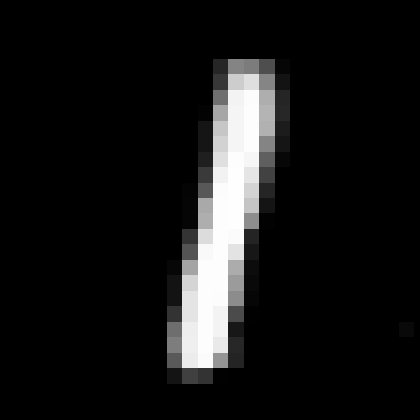} \vspace{0.25cm}
\end{subfigure}
\begin{subfigure}[t]{.12\textwidth}
  \centering
  \includegraphics[width=.8\linewidth]{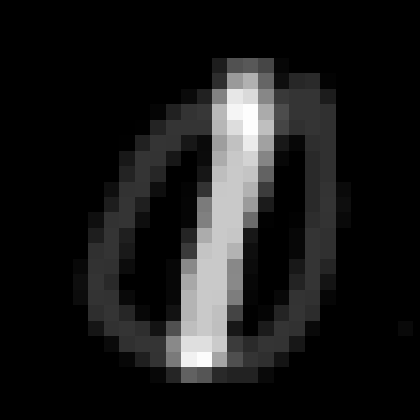} \vspace{0.25cm}
\end{subfigure}
\begin{subfigure}[t]{.12\textwidth}
  \centering
  \includegraphics[width=.8\linewidth]{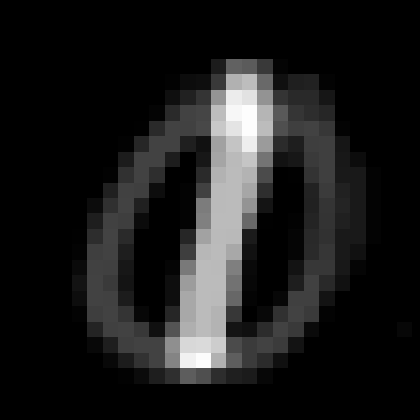} \vspace{0.25cm}
\end{subfigure}
\begin{subfigure}[t]{.12\textwidth}
  \centering
  \includegraphics[width=.8\linewidth]{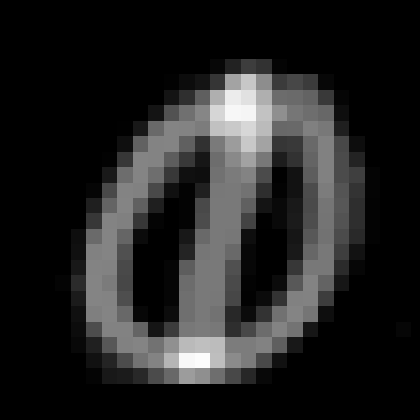} \vspace{0.25cm}
\end{subfigure}
\begin{subfigure}[t]{.12\textwidth}
  \centering
  \includegraphics[width=.8\linewidth]{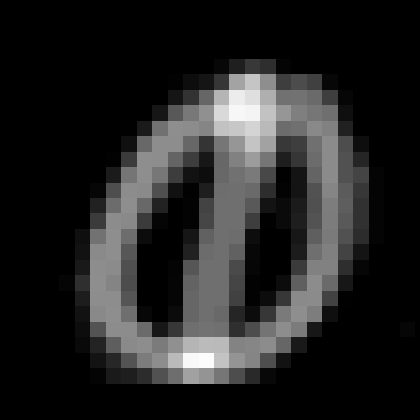} \vspace{0.25cm}
\end{subfigure}
\begin{subfigure}[t]{.12\textwidth}
  \centering
  \includegraphics[width=.8\linewidth]{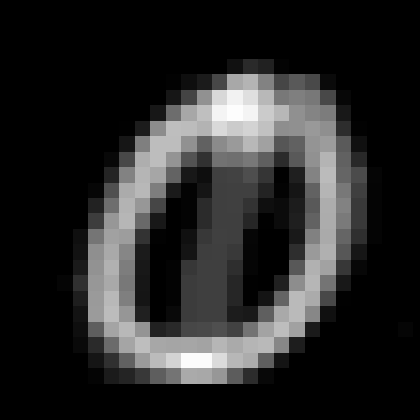} \vspace{0.25cm}
\end{subfigure}
\begin{subfigure}[t]{.12\textwidth}
  \centering
  \includegraphics[width=.8\linewidth]{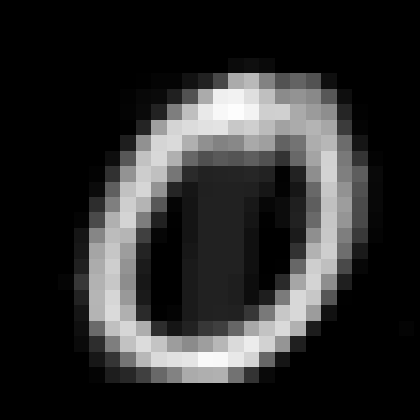} \vspace{0.25cm}
\end{subfigure}

\begin{subfigure}[t]{.12\textwidth}
  \centering
  \includegraphics[width=.8\linewidth]{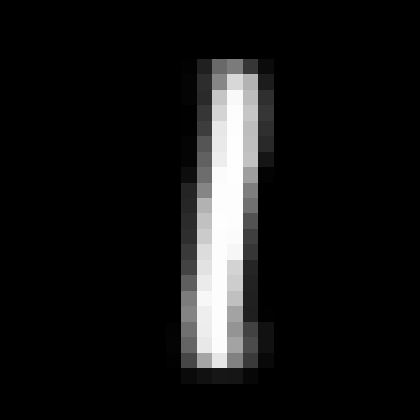} \vspace{0.25cm}
\end{subfigure}
\begin{subfigure}[t]{.12\textwidth}
  \centering
  \includegraphics[width=.8\linewidth]{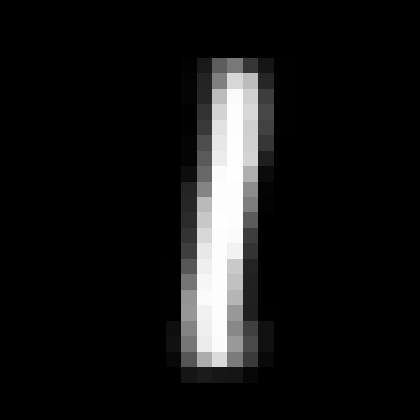} \vspace{0.25cm}
\end{subfigure}
\begin{subfigure}[t]{.12\textwidth}
  \centering
  \includegraphics[width=.8\linewidth]{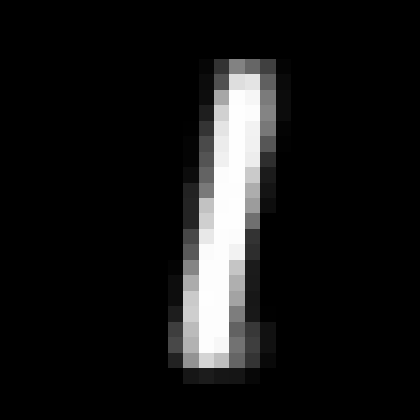} \vspace{0.25cm}
\end{subfigure}
\begin{subfigure}[t]{.12\textwidth}
  \centering
  \includegraphics[width=.8\linewidth]{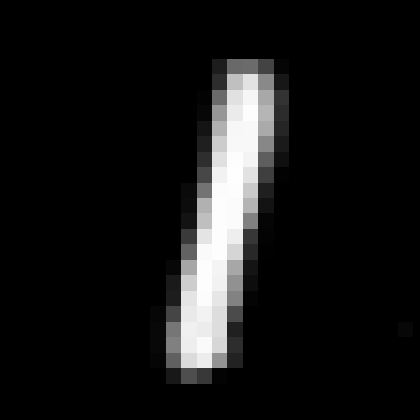} \vspace{0.25cm}
\end{subfigure}
\begin{subfigure}[t]{.12\textwidth}
  \centering
  \includegraphics[width=.8\linewidth]{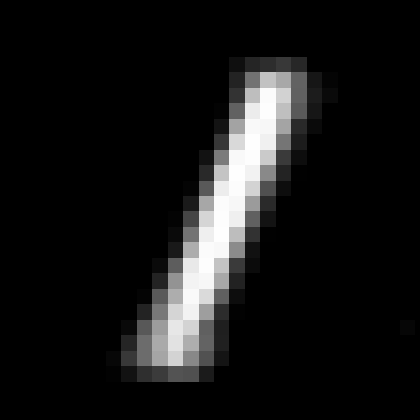} \vspace{0.25cm}
\end{subfigure}
\begin{subfigure}[t]{.12\textwidth}
  \centering
  \includegraphics[width=.8\linewidth]{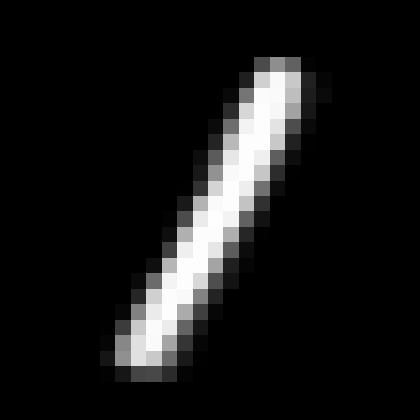} \vspace{0.25cm}
\end{subfigure}
\begin{subfigure}[t]{.12\textwidth}
  \centering
  \includegraphics[width=.8\linewidth]{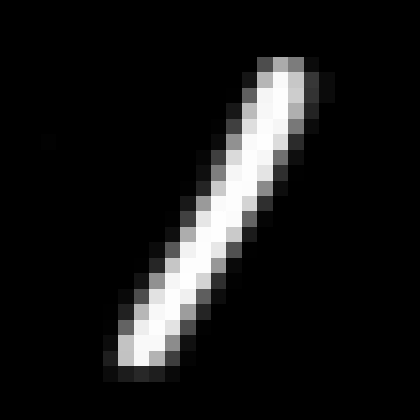} \vspace{0.25cm}
\end{subfigure} 

\begin{subfigure}[t]{.12\textwidth}
  \centering
  \includegraphics[width=.8\linewidth]{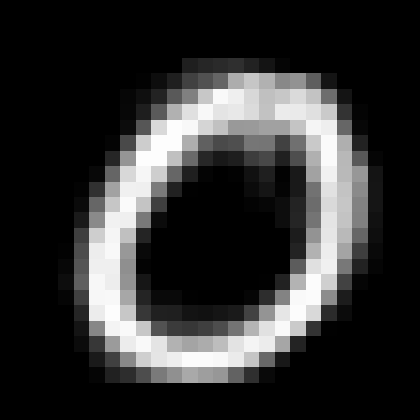} \vspace{0.25cm}
\end{subfigure}
\begin{subfigure}[t]{.12\textwidth}
  \centering
  \includegraphics[width=.8\linewidth]{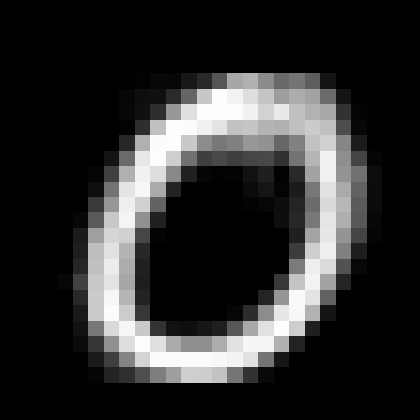} \vspace{0.25cm}
\end{subfigure}
\begin{subfigure}[t]{.12\textwidth}
  \centering
  \includegraphics[width=.8\linewidth]{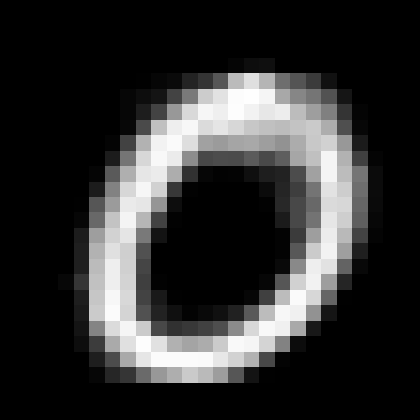} \vspace{0.25cm}
\end{subfigure}
\begin{subfigure}[t]{.12\textwidth}
  \centering
  \includegraphics[width=.8\linewidth]{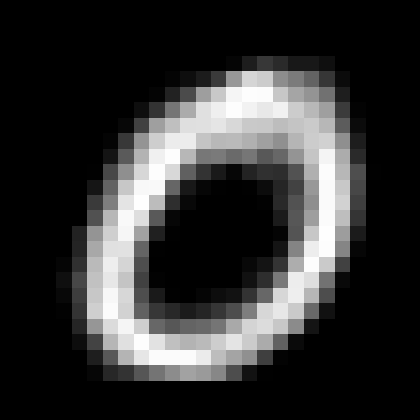} \vspace{0.25cm}
\end{subfigure}
\begin{subfigure}[t]{.12\textwidth}
  \centering
  \includegraphics[width=.8\linewidth]{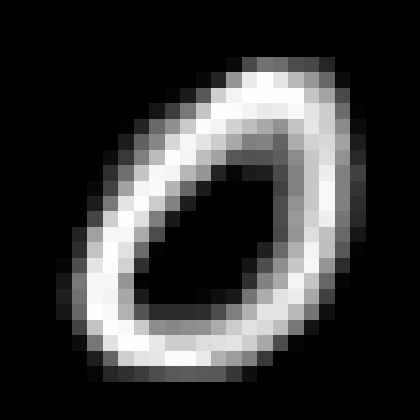} \vspace{0.25cm}
\end{subfigure}
\begin{subfigure}[t]{.12\textwidth}
  \centering
  \includegraphics[width=.8\linewidth]{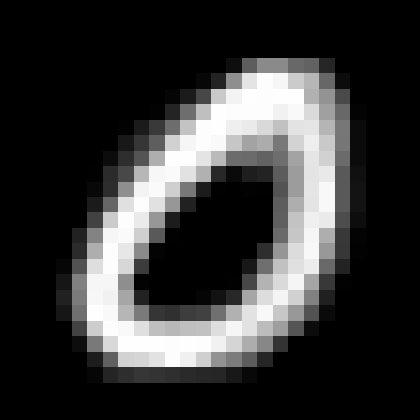} \vspace{0.25cm}
\end{subfigure}
\begin{subfigure}[t]{.12\textwidth}
  \centering
  \includegraphics[width=.8\linewidth]{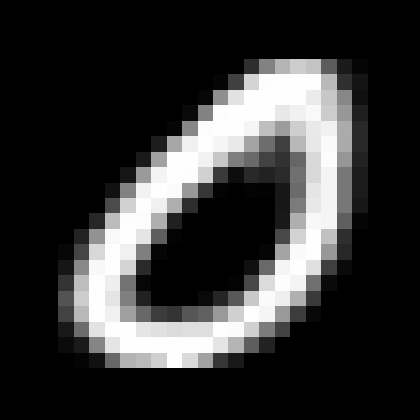} \vspace{0.25cm}
\end{subfigure}
 \caption{Exploration of the latent space of the MNIST digits data set. In the top two rows images are generated along the directions A and B in Fig.\ref{fig:MNIST_feat_space_12} and in the bottom row the images are generated along direction C in Fig. \ref{fig:MNIST_feat_space_13}.}
    \label{fig:exploring_MNIST_directions}
\end{figure}

In Fig. \ref{fig:exploring_MNIST_directions}, digits are generated along the directions indicated on the plots of the latent space in Fig. \ref{fig:MNIST_exploration}. This allows us to interpret the different regions and the meaning of the principal components. Along direction A, corresponding to the first principal component, we find an interpolation between the regions with digits of zero and one. Direction B seems to correlate with the orientation of the digit. This explains the smaller variation along the second principal component for the zeros as rotating the digit zero has a smaller effect compared to the rotation of digit one. The third direction, corresponding to component 3, seems to be related to squeezing the zeros together, which explains the larger variance for the zeros compared to the ones. 
\newpage
\subsubsection{Yale Face Database}\hfill

Another example of latent space exploration is done on the Extended Yale Face Database B~\cite{GeBeKr01}, where 1720 data points are sampled. A Gaussian kernel with bandwidth $\sigma^2 = 650$, $S = 45$ and number of components $d=20$.

\begin{figure}[h]
\begin{subfigure}{0.4\textwidth}
\begin{subfigure}[t]{0.49\textwidth}
    \centering
  \includegraphics[width=.9\linewidth]{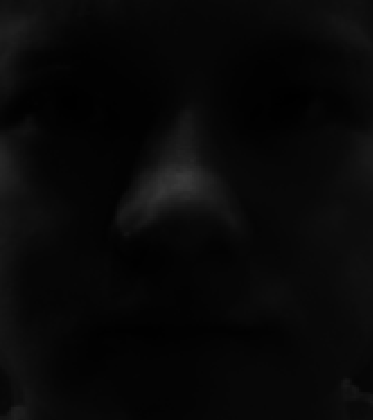}
  \caption{}
\end{subfigure}\hfill
\begin{subfigure}[t]{0.49\textwidth}
    \centering
  \includegraphics[width=.9\linewidth]{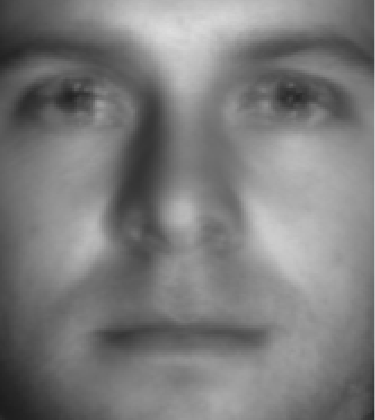}
  \caption{}
\end{subfigure}\hfill
\begin{subfigure}[t]{0.49\textwidth}
    \centering
  \includegraphics[width=.9\linewidth]{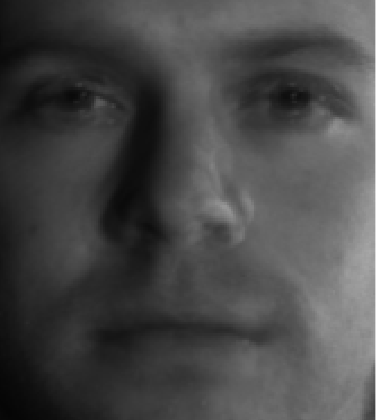}
  \caption{}
\end{subfigure}\hfill
\begin{subfigure}[t]{0.49\textwidth}
    \centering
  \includegraphics[width=.9\linewidth]{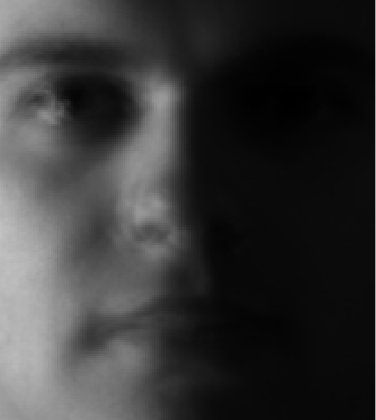}
  \caption{}
\end{subfigure}\hfill
\label{fig:gen_faces_regions}
\end{subfigure}\hfill
\begin{subfigure}{0.6\textwidth}
  \centering
    \includegraphics[width=1.1\linewidth]{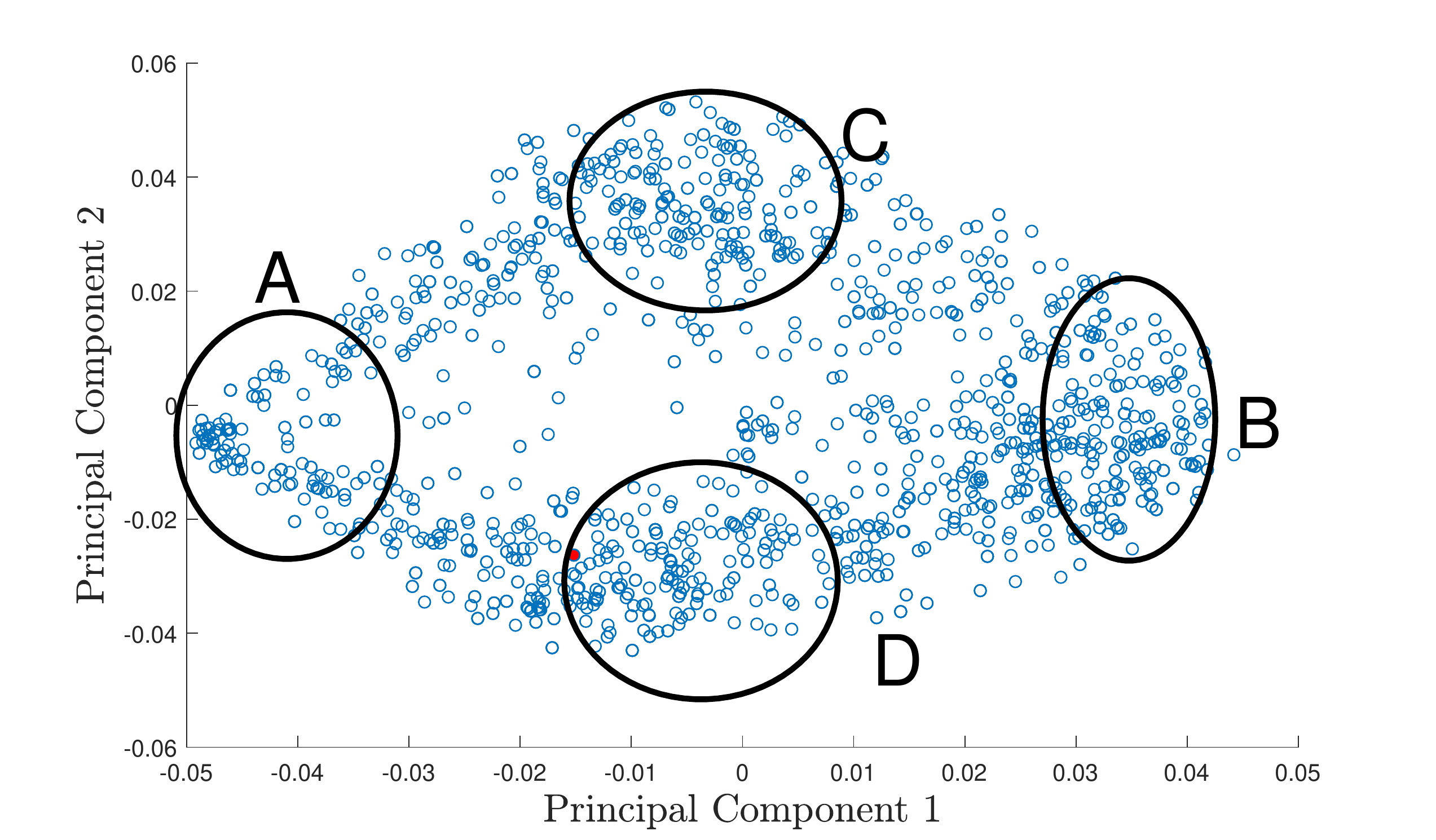}
    \caption{}
    \label{fig:feature_space_yale}
\end{subfigure}\hfill
\caption{Exploring different regions of the latent space of the Yale Face Database. (e) Data projected on the first two principal components for the Yale Face Database. (a)-(d) Generated faces from the different regions.}
\label{fig:latent_space_exploration_yale}
\end{figure}

The latent space along the first two principal components is shown in Fig. \ref{fig:feature_space_yale}. Four different regions within the feature space are highlighted from which corresponding images are generated. The dissimilarity between the images in the various regions suggests the components capture different lighting conditions on the subjects.

\begin{figure}[h]
\centering
\begin{subfigure}{0.45\textwidth}
\centering
    \includegraphics[width=1\linewidth]{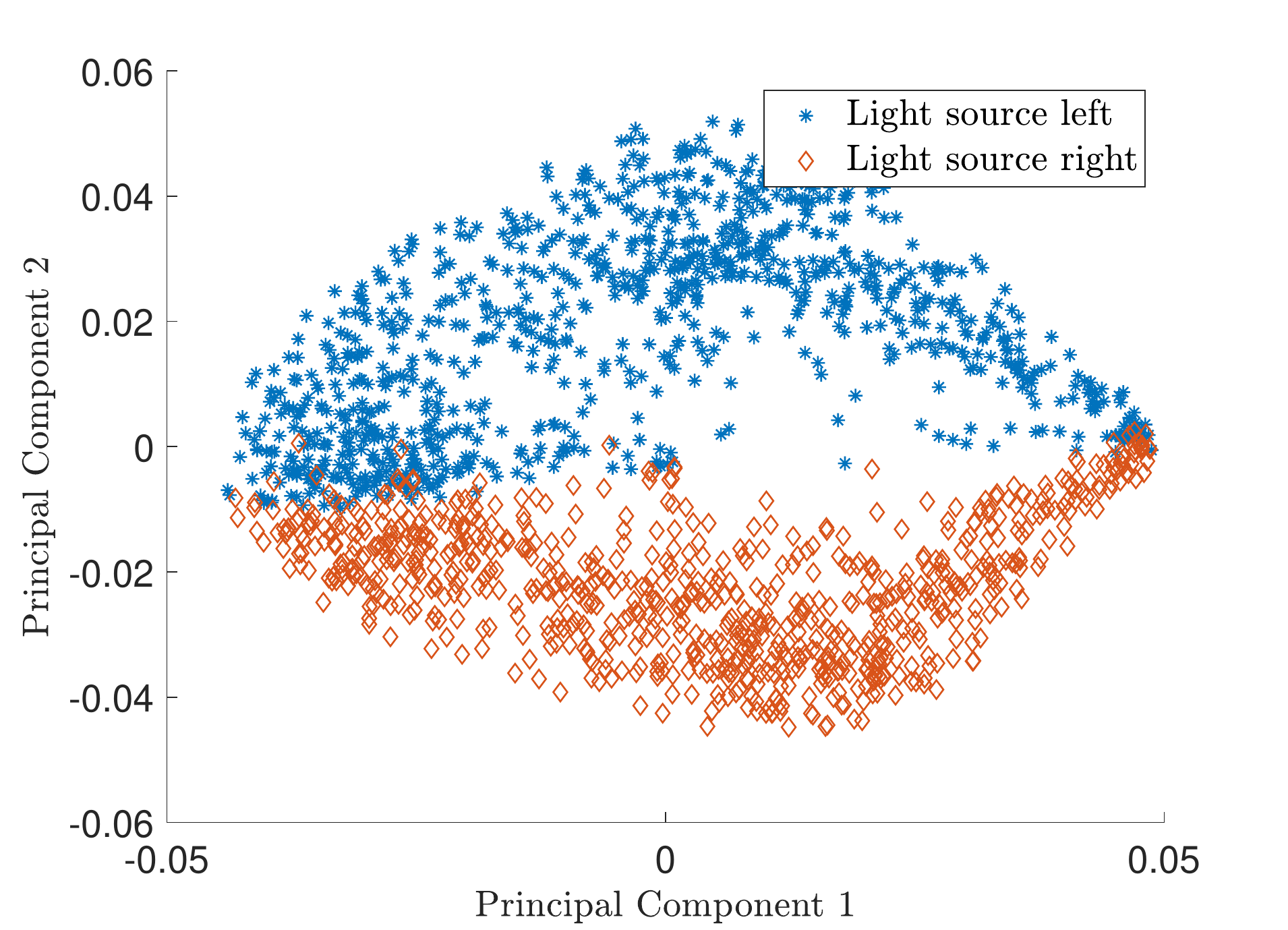}
    \caption{}
    \label{fig:lightsource}
\end{subfigure}
\begin{subfigure}{0.45\textwidth}
\centering
    \includegraphics[width=1\linewidth]{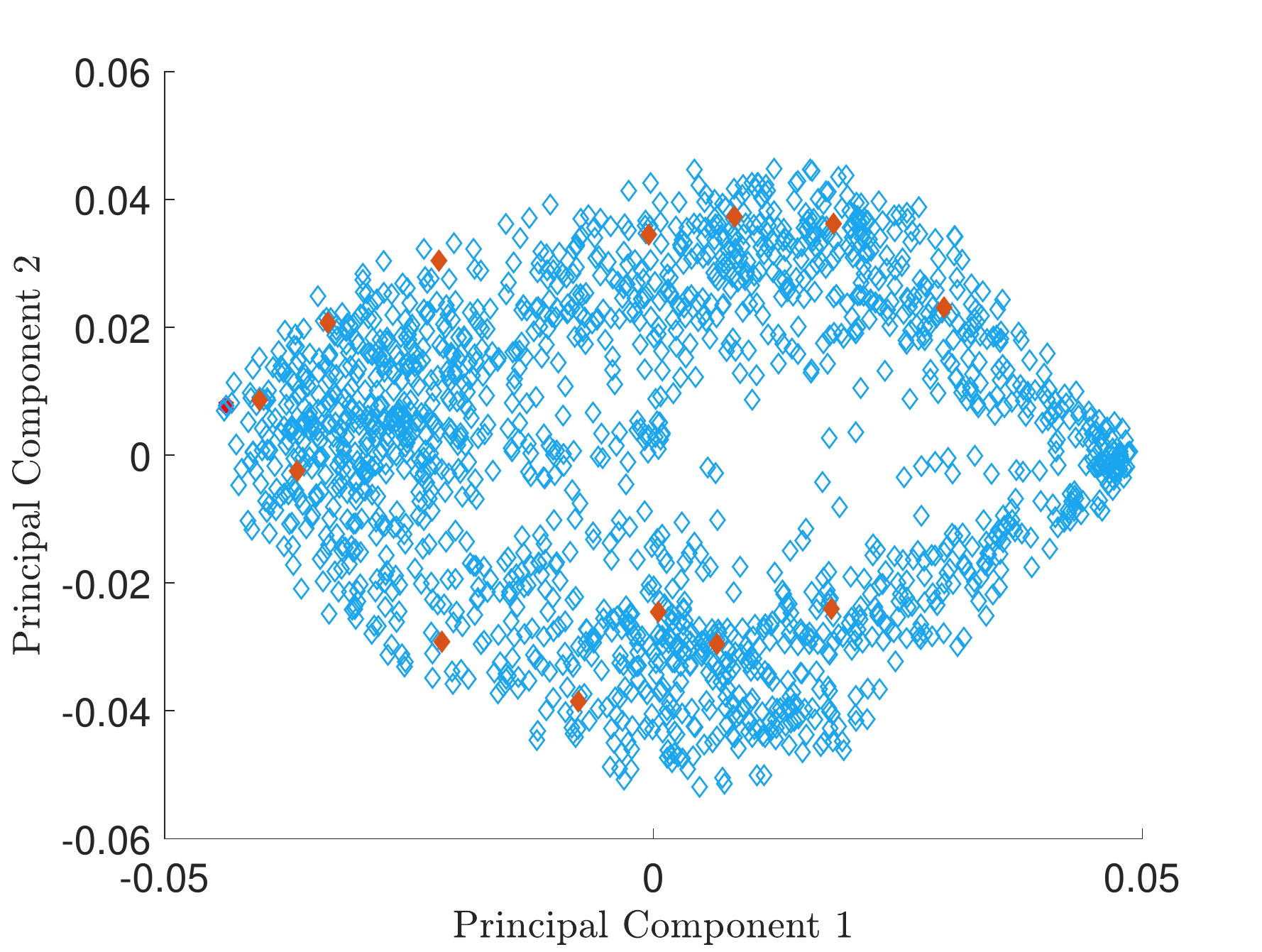}
    \caption{}
    \label{fig:testset}
\end{subfigure}
 \caption{Latent space of the Yale Facebase database B. (a) Points in orange indicate data points with a negative azimuthal angle between the camera direction and source of illumination, which corresponds to a light source to the right of the subject and vice versa for positive azimuthal angle. (b) Points in red indicate the hidden units of the same subject with lighting from different azimuthal angles.}
    \label{fig:validationYaleface}
\end{figure}

\begin{figure}[H]
\begin{subfigure}{.24\textwidth}
  \centering
  \includegraphics[width=.6\linewidth]{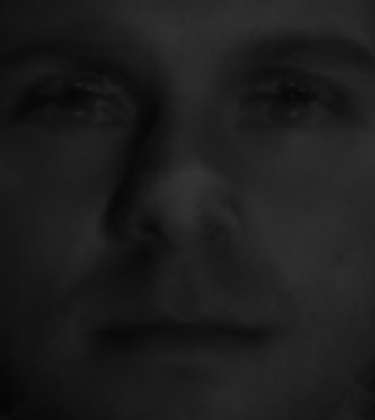}
\end{subfigure}
\begin{subfigure}{.24\textwidth}
  \centering
  \includegraphics[width=.6\linewidth]{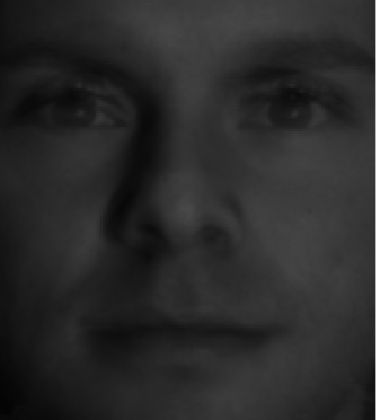}
\end{subfigure}
\begin{subfigure}{.24\textwidth}
  \centering
  \includegraphics[width=.6\linewidth]{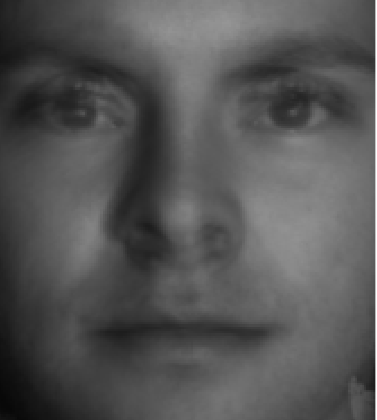}
\end{subfigure}
\begin{subfigure}{.24\textwidth}
  \centering
  \includegraphics[width=.6\linewidth]{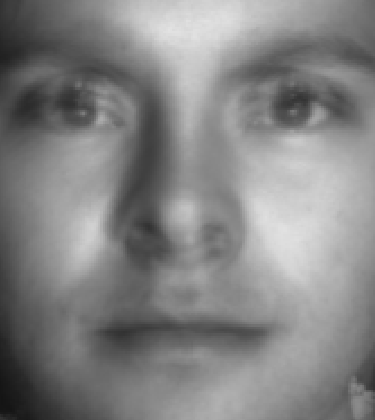}
\end{subfigure}\vspace{8pt}
\begin{subfigure}{.24\textwidth}
  \centering
  \includegraphics[width=.6\linewidth]{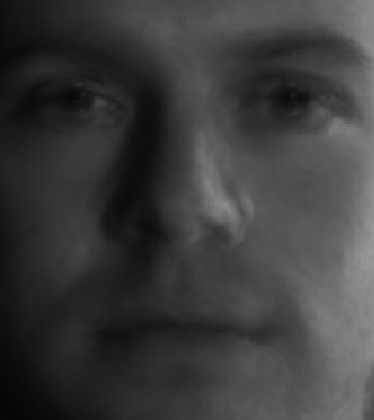}
\end{subfigure}\hfill
\begin{subfigure}{.24\textwidth}
  \centering
  \includegraphics[width=.6\linewidth]{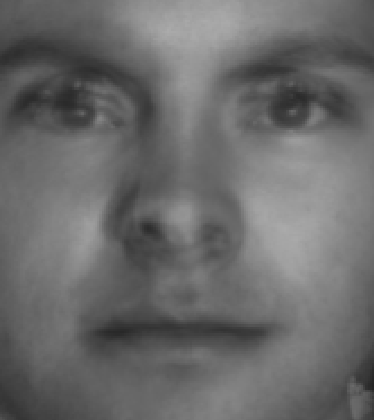}
\end{subfigure}\hfill
\begin{subfigure}{.24\textwidth}
  \centering
  \includegraphics[width=.6\linewidth]{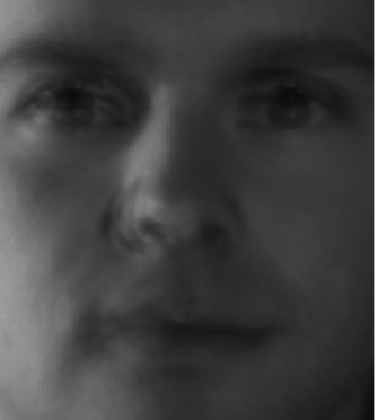}
\end{subfigure}\hfill
\begin{subfigure}{.24\textwidth}
  \centering
  \includegraphics[width=.6\linewidth]{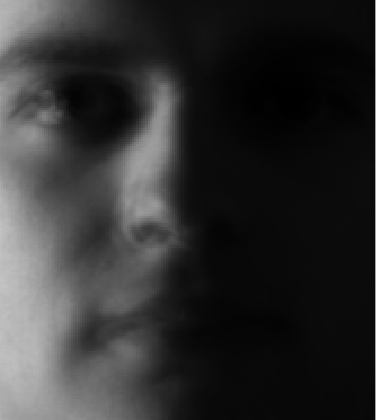}
\end{subfigure}
\caption{Exploring the space between the regions of the latent space in figure \ref{fig:feature_space_yale}.The top row shows images generated between regions A and B, while the bottom row explores the space between regions C and D.}
\label{fig:interpolation_regions_yale}
\end{figure}

The tool allows us to gradually move between these different regions and see the changes in the input space as shown in Fig. \ref{fig:interpolation_regions_yale}. Moving between regions A and B shows increasing illumination of the subject. We can thus interpret the first principal component as determining the global level of illumination. Note that besides data points without a light source no variation of the intensity of the lighting was varied while collecting the data for the Yale Face Database B. Only the position of the light source was changed. Generative Kernel PCA thus allows us to control the level of illumination regardless of the position of the light source. The bottom row seems to indicate that the second principal component can be interpreted as the position of the light source. In region C of the feature space the points are illuminated from the right and region D from the left.This interpretation of the second principal component seems indeed valid from Fig. \ref{fig:lightsource} where the latent space is visualised with labels indicating the position of illumination obtained from the Yale Face Database. Faces with a positive azimuthal angle between the camera direction and the source of illumination are contained in the top half of the figure. This corresponds to a light source left of the subject and vice versa for a negative azimuthal angle. The first and second component are thus disentangled as the level of illumination does not determine whether the light comes from the left or the right. Furthermore in Fig. \ref{fig:testset} the hidden units corresponding to the same subject under different lighting conditions are shown. The elevation of the light source is kept constant at zero elevation, while the azimuthal angle is varied. We see from the plot that the points move not strictly along the second principal component but follow a more circular path. This indicates that varying the azimuthal angle correlates with both the first and second principal component, i.e. moving the light source more to the side also decreases the global illumination level as less light is able to illuminate the face. We conclude that while in the original data set the position of the light source and the level of illumination are correlated, kernel PCA allows us to disentangle these factors and vary them separately when generating new images.

\begin{figure}[h]
\begin{subfigure}{.19\textwidth}
  \centering
  \includegraphics[width=.8\linewidth]{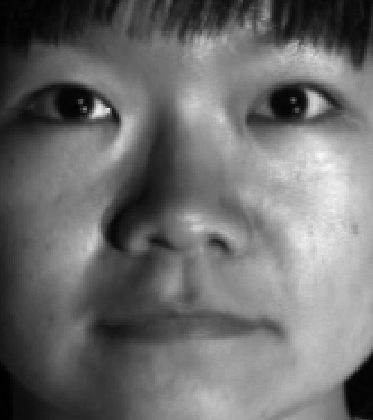}
\end{subfigure}
\begin{subfigure}{.19\textwidth}
  \centering
  \includegraphics[width=.8\linewidth]{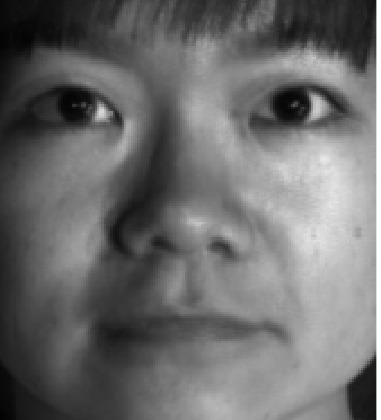}
\end{subfigure}
\begin{subfigure}{.19\textwidth}
  \centering
  \includegraphics[width=.8\linewidth]{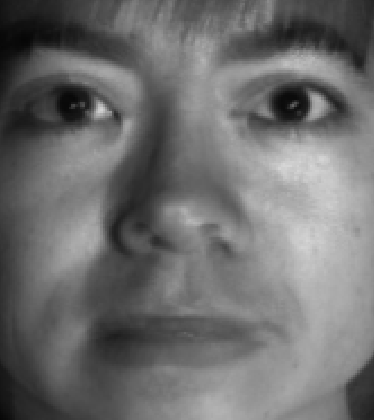}
\end{subfigure}
\begin{subfigure}{.19\textwidth}
  \centering
  \includegraphics[width=.8\linewidth]{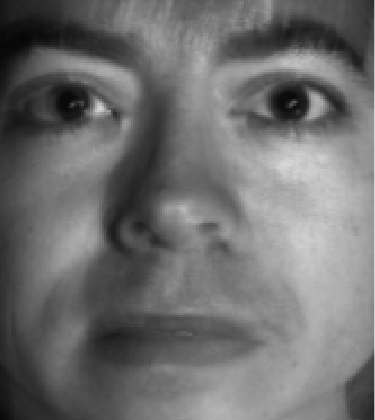}
\end{subfigure}
\begin{subfigure}{.19\textwidth}
  \centering
  \includegraphics[width=.8\linewidth]{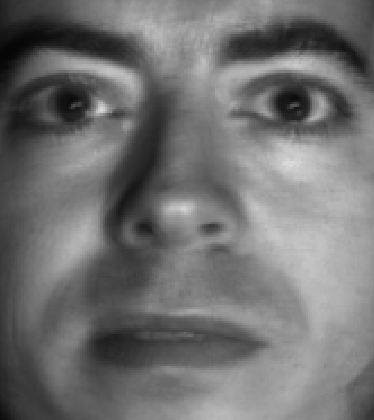}
\end{subfigure}\vspace{8pt}
\begin{subfigure}{.19\textwidth}
  \centering
  \includegraphics[width=.8\linewidth]{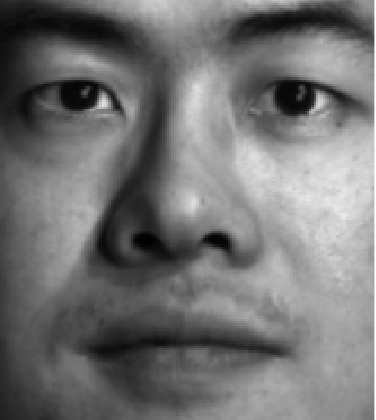}
\end{subfigure}\hfill
\begin{subfigure}{.19\textwidth}
  \centering
  \includegraphics[width=.8\linewidth]{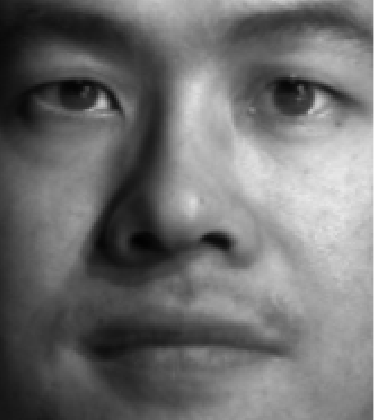}
\end{subfigure}\hfill
\begin{subfigure}{.19\textwidth}
  \centering
  \includegraphics[width=.8\linewidth]{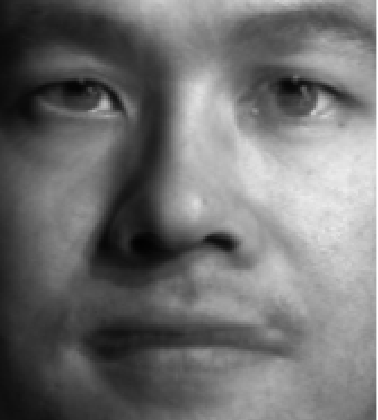}
\end{subfigure}\hfill
\begin{subfigure}{.19\textwidth}
  \centering
  \includegraphics[width=.8\linewidth]{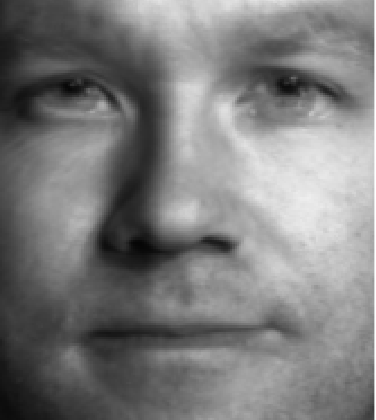}
\end{subfigure}\hfill
\begin{subfigure}{.19\textwidth}
  \centering
  \includegraphics[width=.8\linewidth]{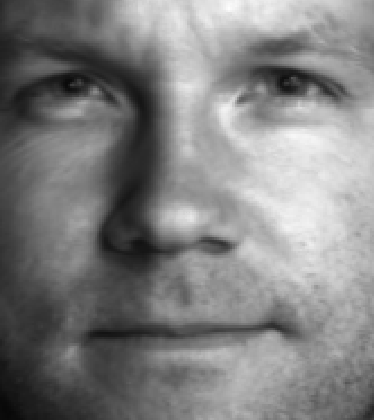}
\end{subfigure}\vspace{8pt}
\begin{subfigure}{.19\textwidth}
  \centering
  \includegraphics[width=.8\linewidth]{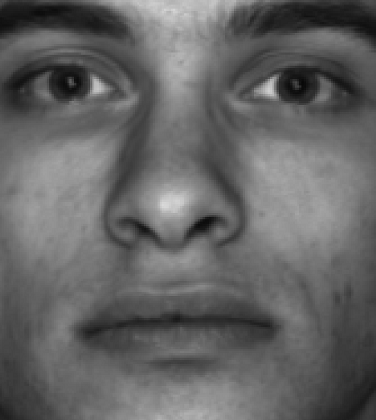}
\end{subfigure}\hfill
\begin{subfigure}{.19\textwidth}
  \centering
  \includegraphics[width=.8\linewidth]{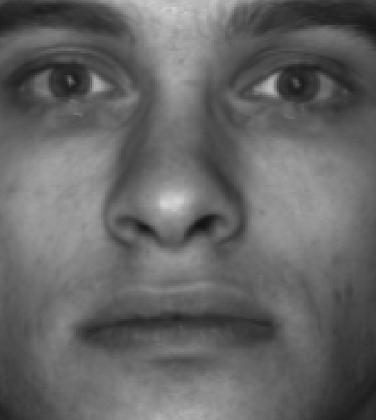}
\end{subfigure}\hfill
\begin{subfigure}{.19\textwidth}
  \centering
  \includegraphics[width=.8\linewidth]{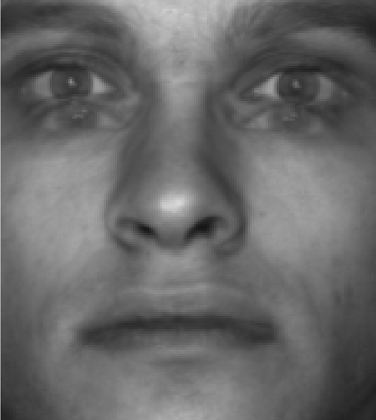}
\end{subfigure}\hfill
\begin{subfigure}{.19\textwidth}
  \centering
  \includegraphics[width=.8\linewidth]{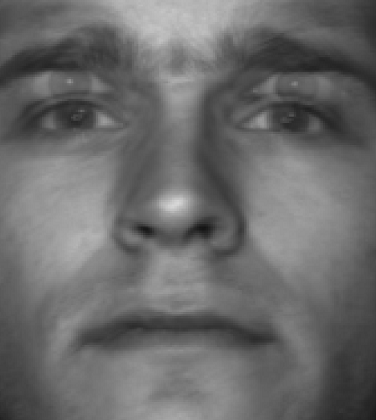}
\end{subfigure}\hfill
\begin{subfigure}{.19\textwidth}
  \centering
  \includegraphics[width=.8\linewidth]{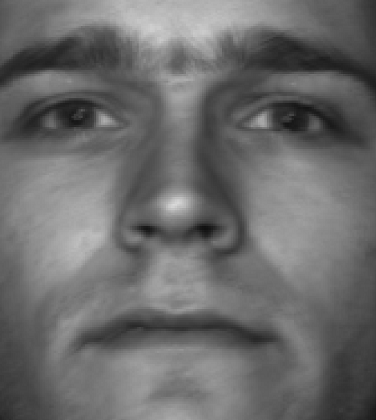}
\end{subfigure}\hfill
\caption{Three examples of interpolation between two subjects of the Yale Face Database B along the fourth component. The uttermost left and right pictures in the rows represent the original faces.}
\label{fig:interpolation_yale}
\end{figure}

As a further example of generative kernel PCA, interpolation between 2 faces is demonstrated. Kernel PCA is performed on a subset of the database consisting out of 130 facial images of two subjects, the hyperparameters are the same as above. Variation along the fourth principal component results in a smooth interpolation between the two subjects, shown in Fig. \ref{fig:interpolation_yale}. We also include an example in the bottom row where the interpolation does not result in a smooth change between the subjects. This illustrates a major limitation of our method as generative kernel PCA predominantly detects global features such as lighting and has difficulty with smaller, local features such as eyes. This stems from the fact that generative kernel PCA relies on the input data to be highly correlated which in this example translates itself to the need of the faces to be aligned with each other. 

\subsubsection{MIT-BIH Arrhythmia database}\hfill

Besides the previous examples of latent space exploration for image datasets, kernel PCA is also applicable to other types of data. In this section, the MIT-BIH Arrhythmia dataset \cite{moody2001impact} is considered consisting out of ECG signals. The goal is to demonstrate the use of kernel PCA to extract interpretable directions in the latent feature space of the ECG signals.This would allow a clinical expert to gain insight and trust in the features extracted by the model. Similar research was previously done by \cite{van2019interpretable} where they investigated the use of disentangled variational autoencoders to extract interpretable ECG embeddings. A similar approach is used to preprocess the data as in \cite{van2019interpretable}. 

The signals from the patients with identifiers 101, 106, 103 and 105 are used for the normal beat signals and the data of patients 102, 104, 107, 217 for the paced beat signals. This results in a total of 785 beat patterns which are processed through a peak detection program \cite{sedghamiz2013online}. The ECG signal is first passed through a fifth-order Butterworth bandpass filter with a lower cutoff frequency of 1Hz and upper cutoff frequency of 60Hz. The ECG beats are sampled at 360Hz and a window of 0.5 seconds is taken around each R-wave resulting in 180 samples per epoch. A regular Gaussian kernel with bandwidth $\sigma^2 = 10$ is used, with $S = 10$. 
The first 10 principal components are used in the reconstruction.

\begin{figure}[H]
\centering
\begin{subfigure}{0.49\textwidth}
    \centering
    \includegraphics[width=1.0\linewidth]{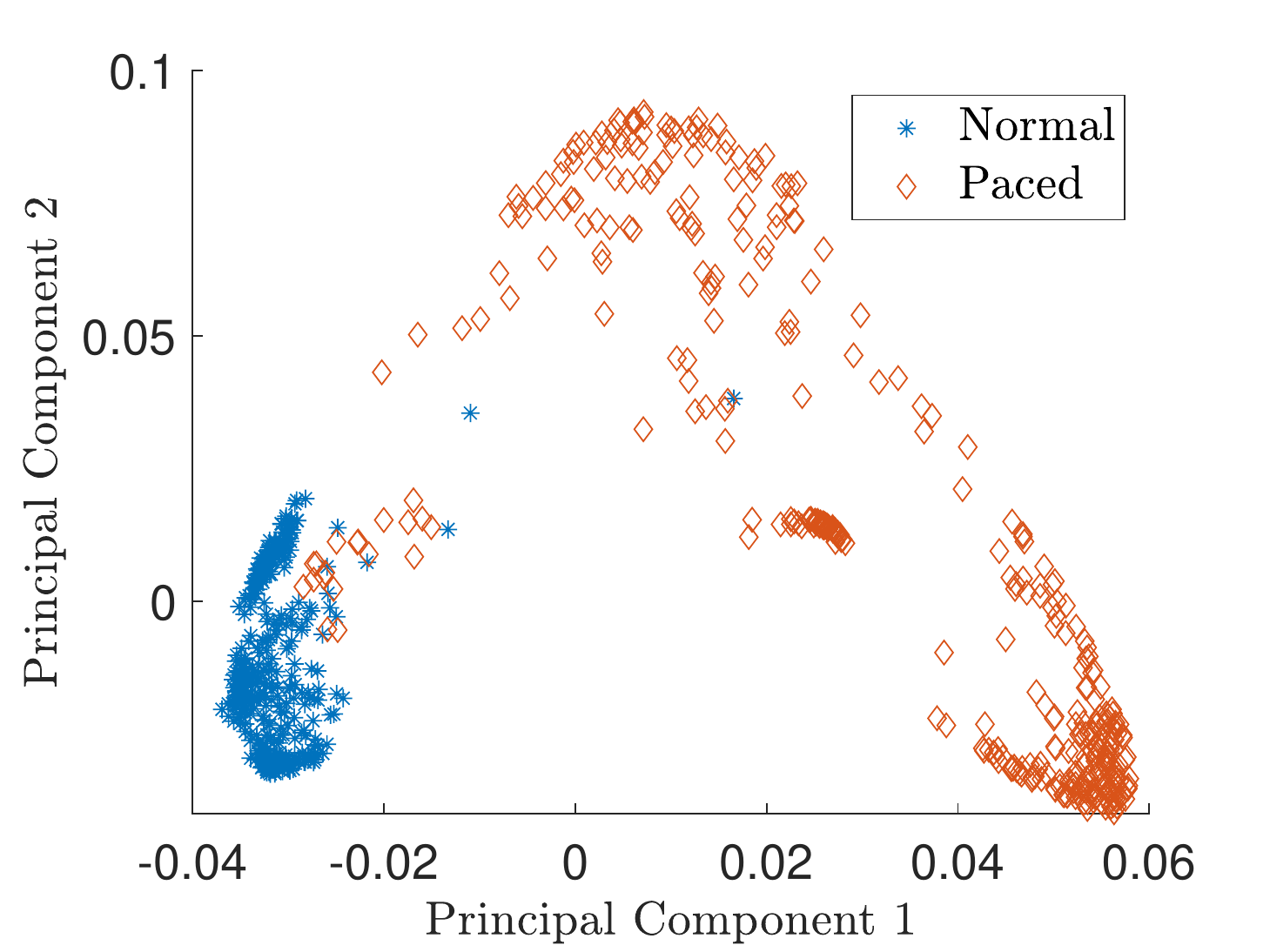}
    \caption{}
    \label{fig:ECG_feat_space_12}
\end{subfigure}
\begin{subfigure}{0.49\textwidth}
\centering
    \includegraphics[width=1.0\linewidth]{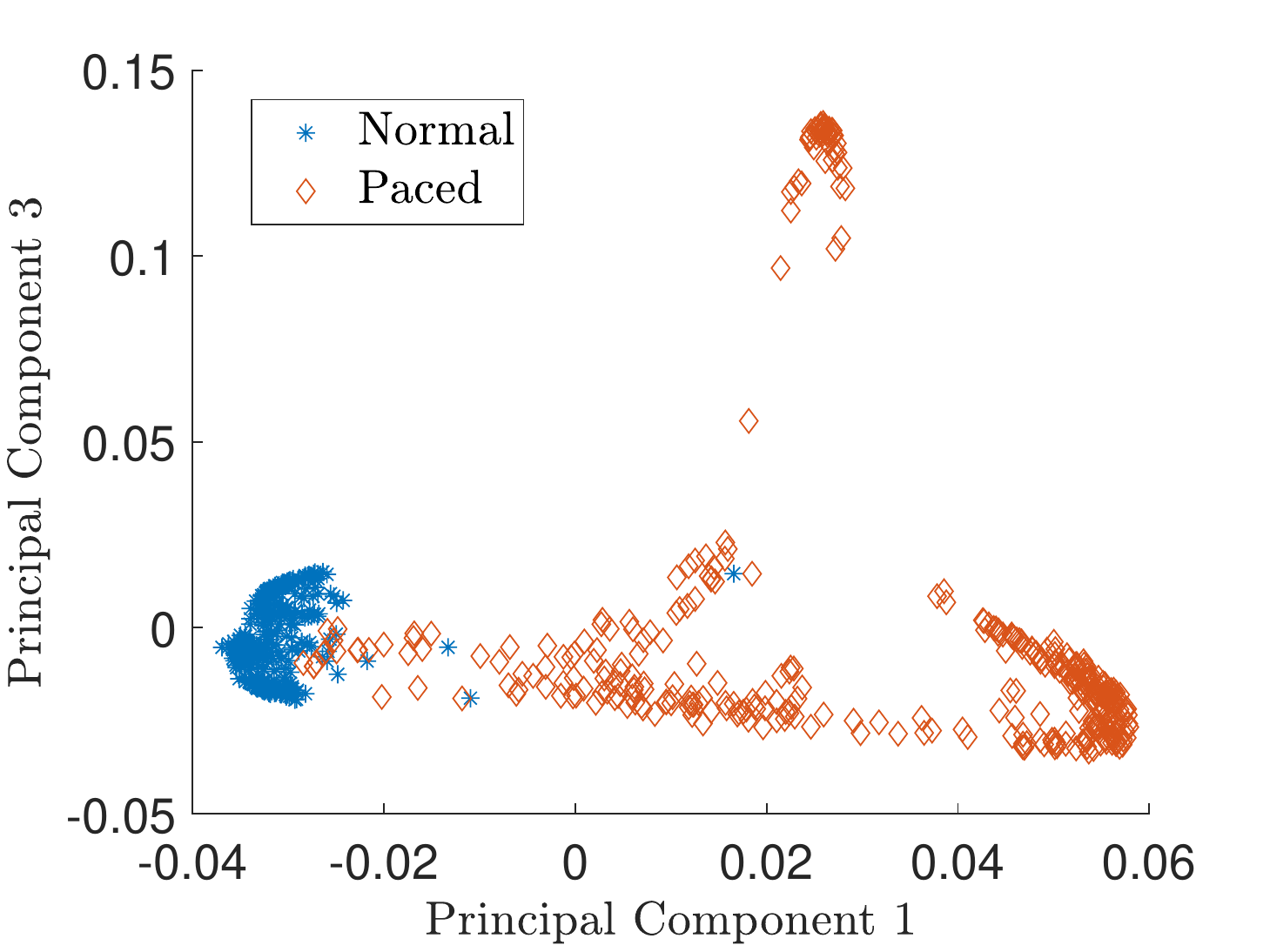}
    \caption{}
    \label{fig:ECG_feat_space_13}
\end{subfigure}
\caption{The latent space for 785 ECG beat signals of the MIT-BIH Arrhythmia dataset projected on different principal components. The hidden units of both normal and paced heartbeats are shown.}
    \label{fig:ECG_feature_space}
\end{figure}

In Fig. \ref{fig:ECG_feature_space} the latent feature space projected on the first two principal components is shown. Kernel PCA is also able to separate between the normal and paced beats. 

\begin{figure}[H]
\begin{subfigure}{.24\textwidth}
  \centering
  \includegraphics[width=1.2\linewidth]{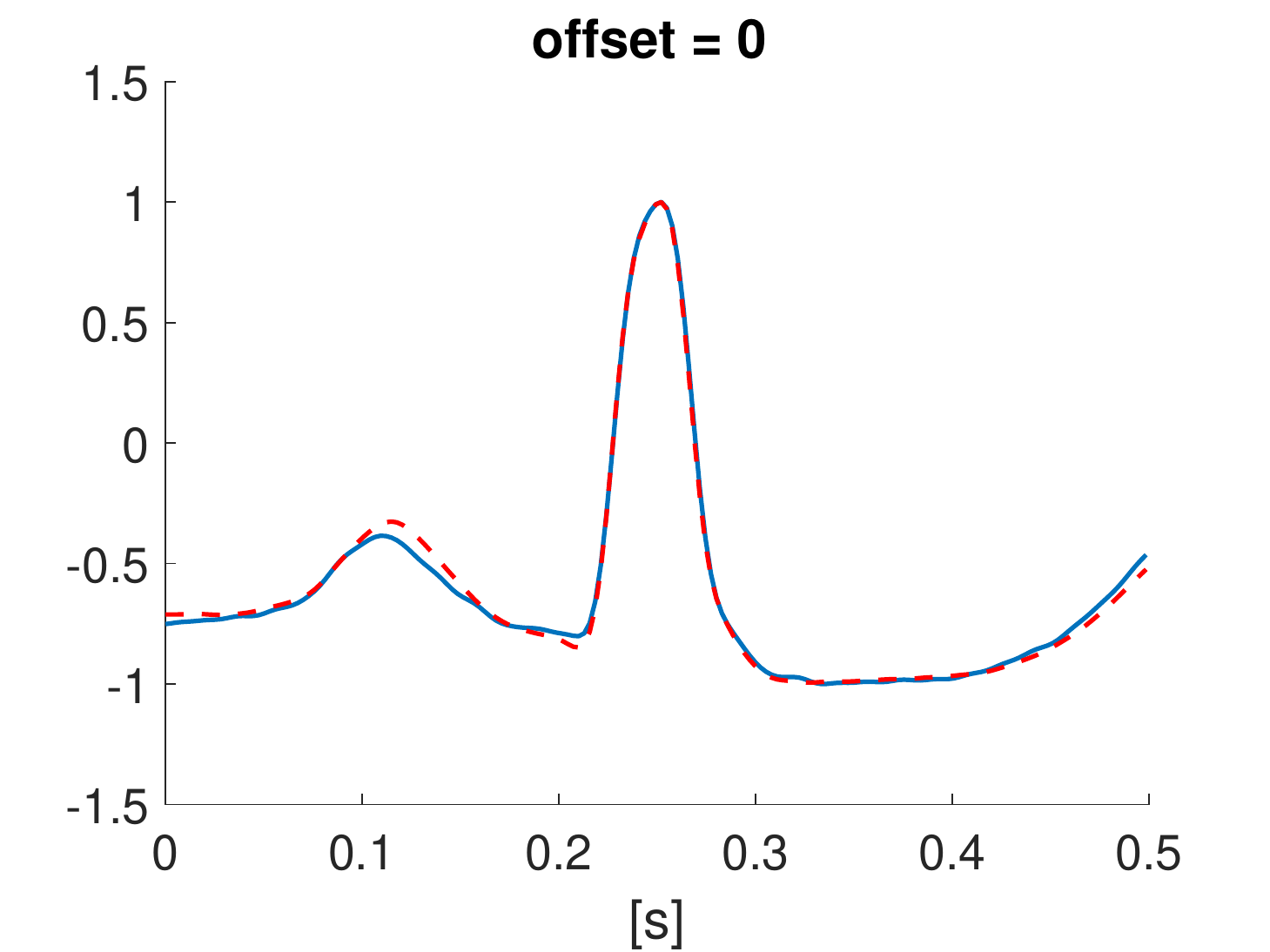}
\end{subfigure}\hfill
\begin{subfigure}{.24\textwidth}
  \centering
  \includegraphics[width=1.2\linewidth]{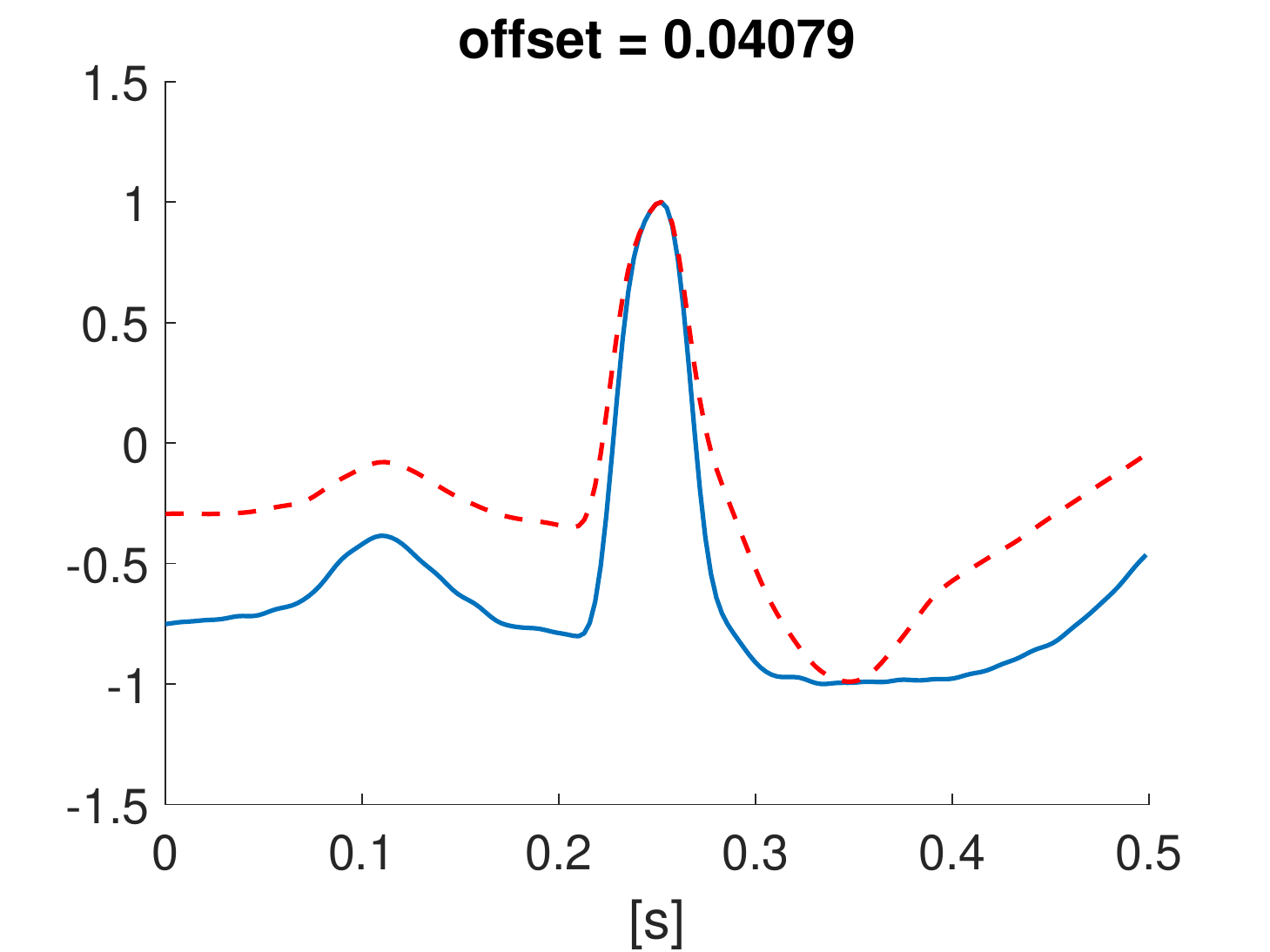}
\end{subfigure}\hfill
\begin{subfigure}{.24\textwidth}
  \centering
  \includegraphics[width=1.2\linewidth]{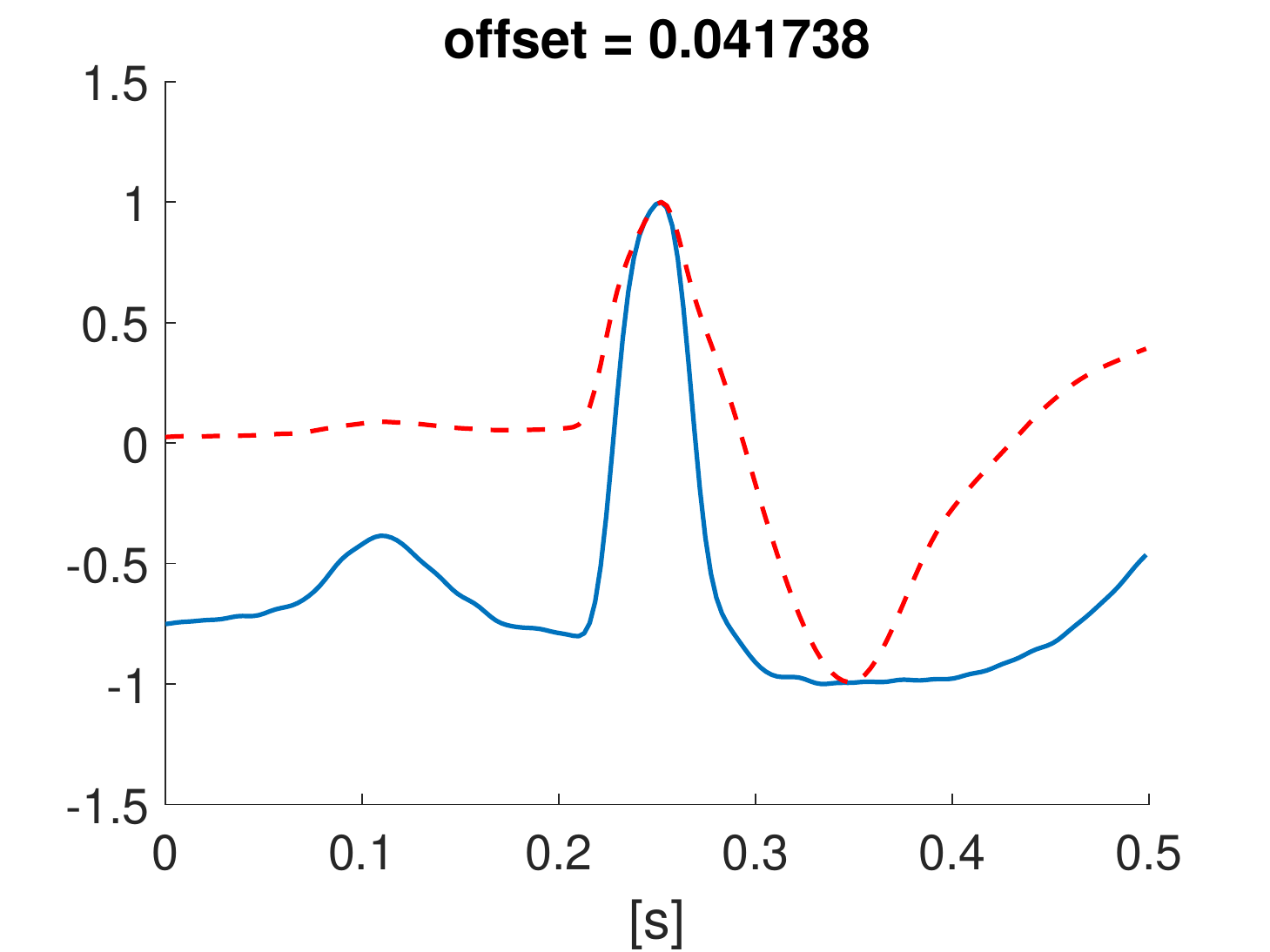}
\end{subfigure}\hfill
\begin{subfigure}{.24\textwidth}
  \centering
  \includegraphics[width=1.2\linewidth]{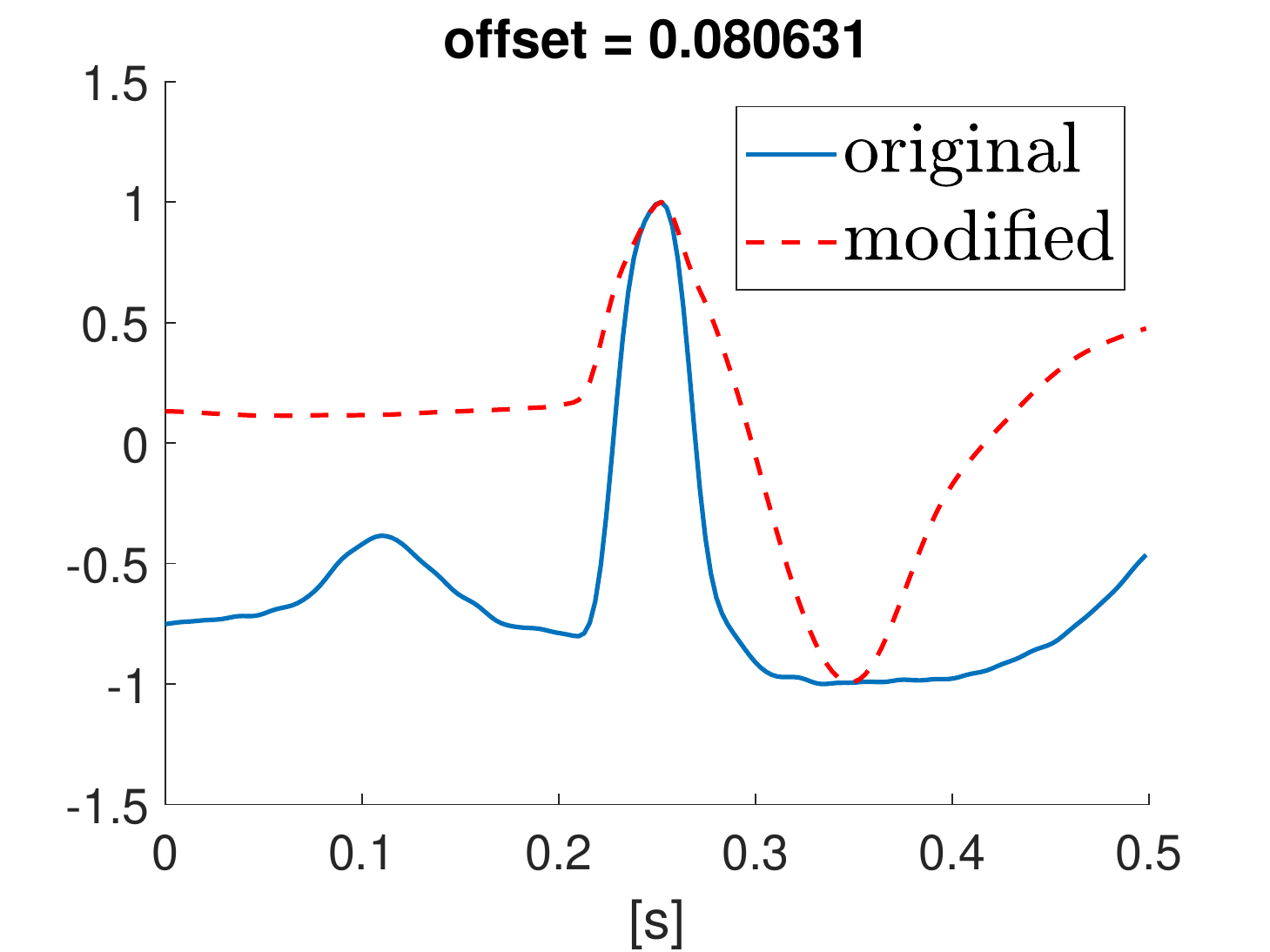}
\end{subfigure}\vspace{4pt}
\begin{subfigure}{.24\textwidth}
  \centering
  \includegraphics[width=1.2\linewidth]{Images/ECG_pc1_normal_1d.pdf}
\end{subfigure}\hfill
\begin{subfigure}{.24\textwidth}
  \centering
  \includegraphics[width=1.2\linewidth]{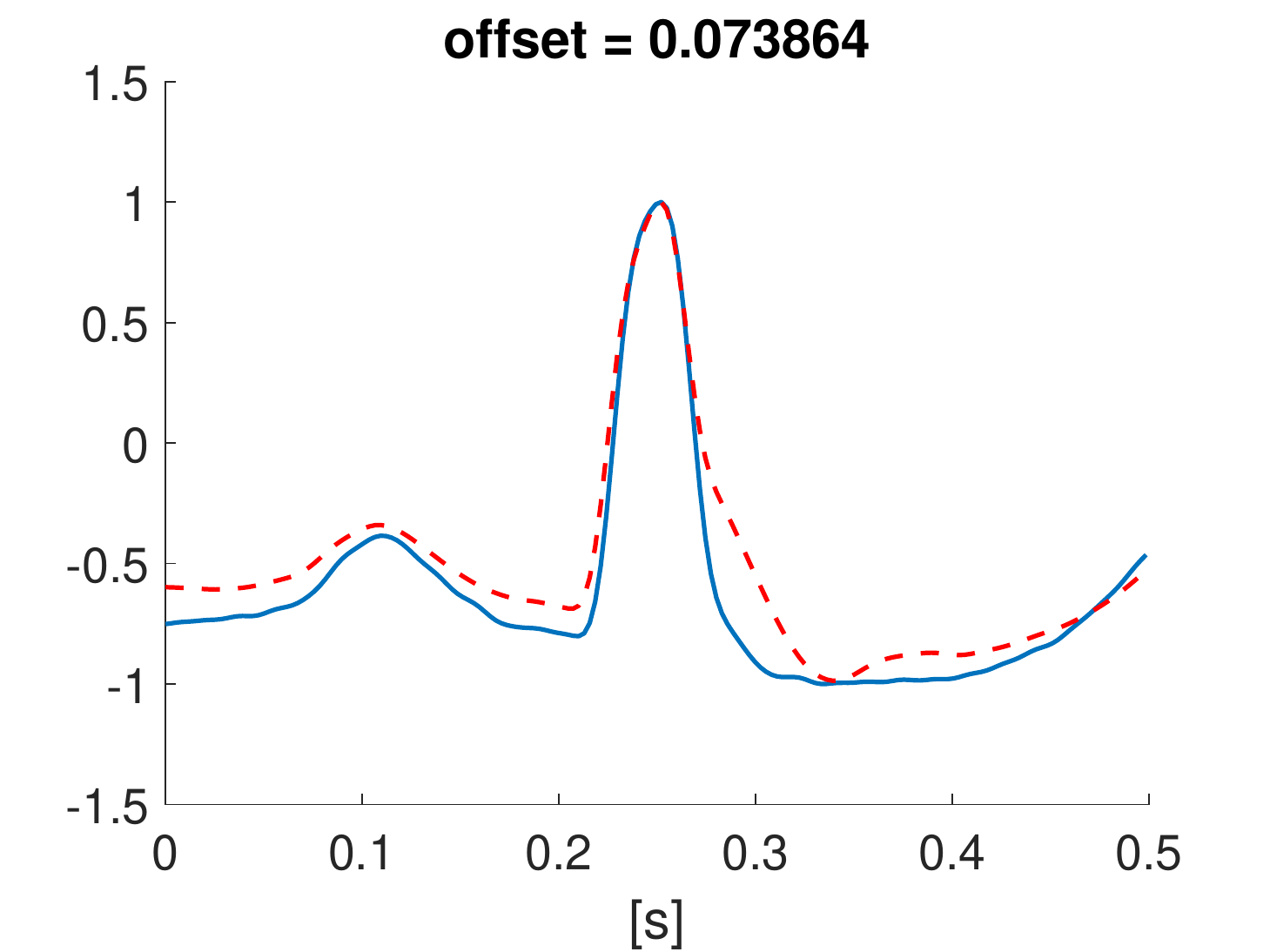}
\end{subfigure}\hfill
\begin{subfigure}{.24\textwidth}
  \centering
  \includegraphics[width=1.2\linewidth]{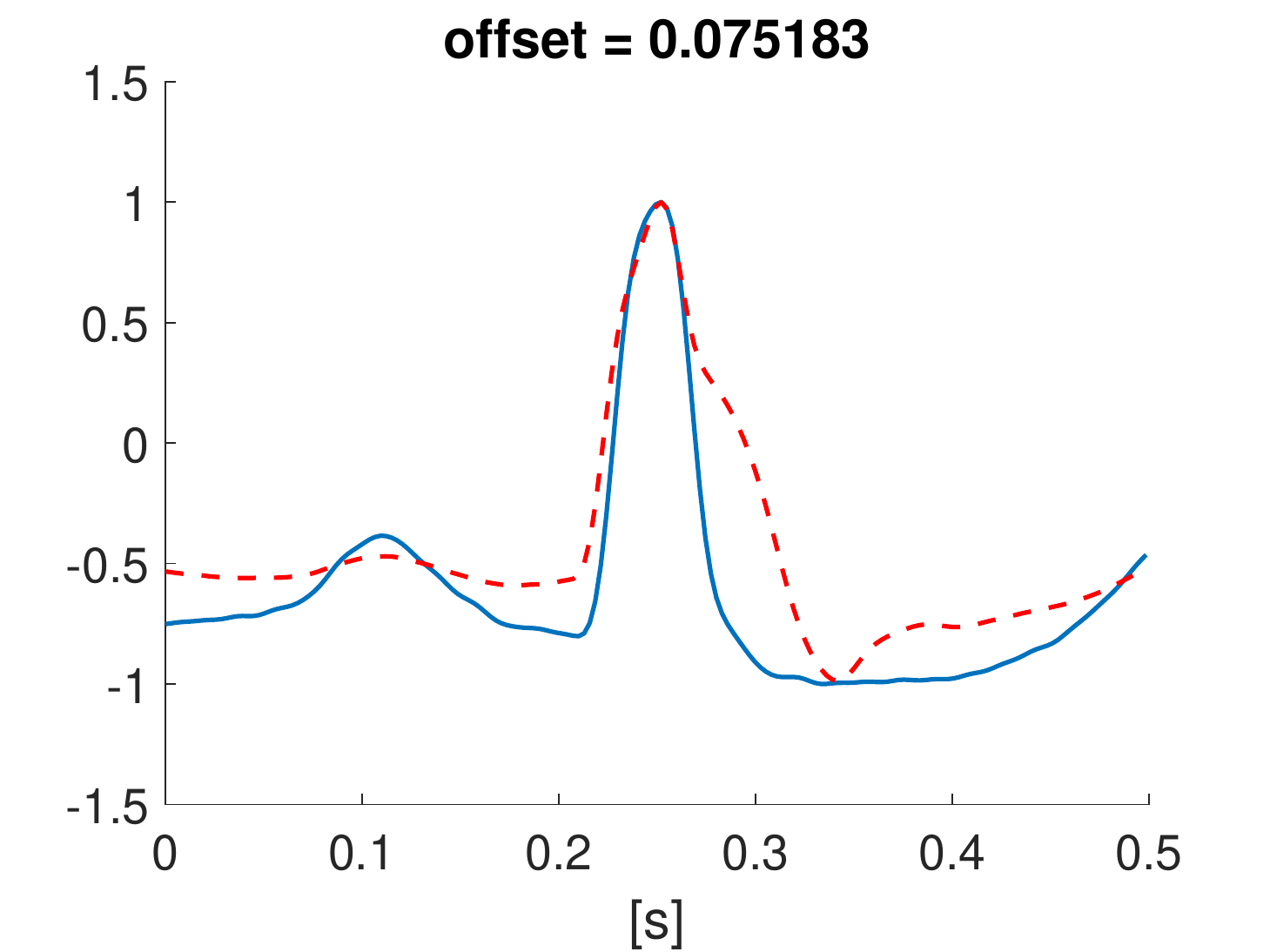}
\end{subfigure}\hfill
\begin{subfigure}{.24\textwidth}
  \centering
  \includegraphics[width=1.2\linewidth]{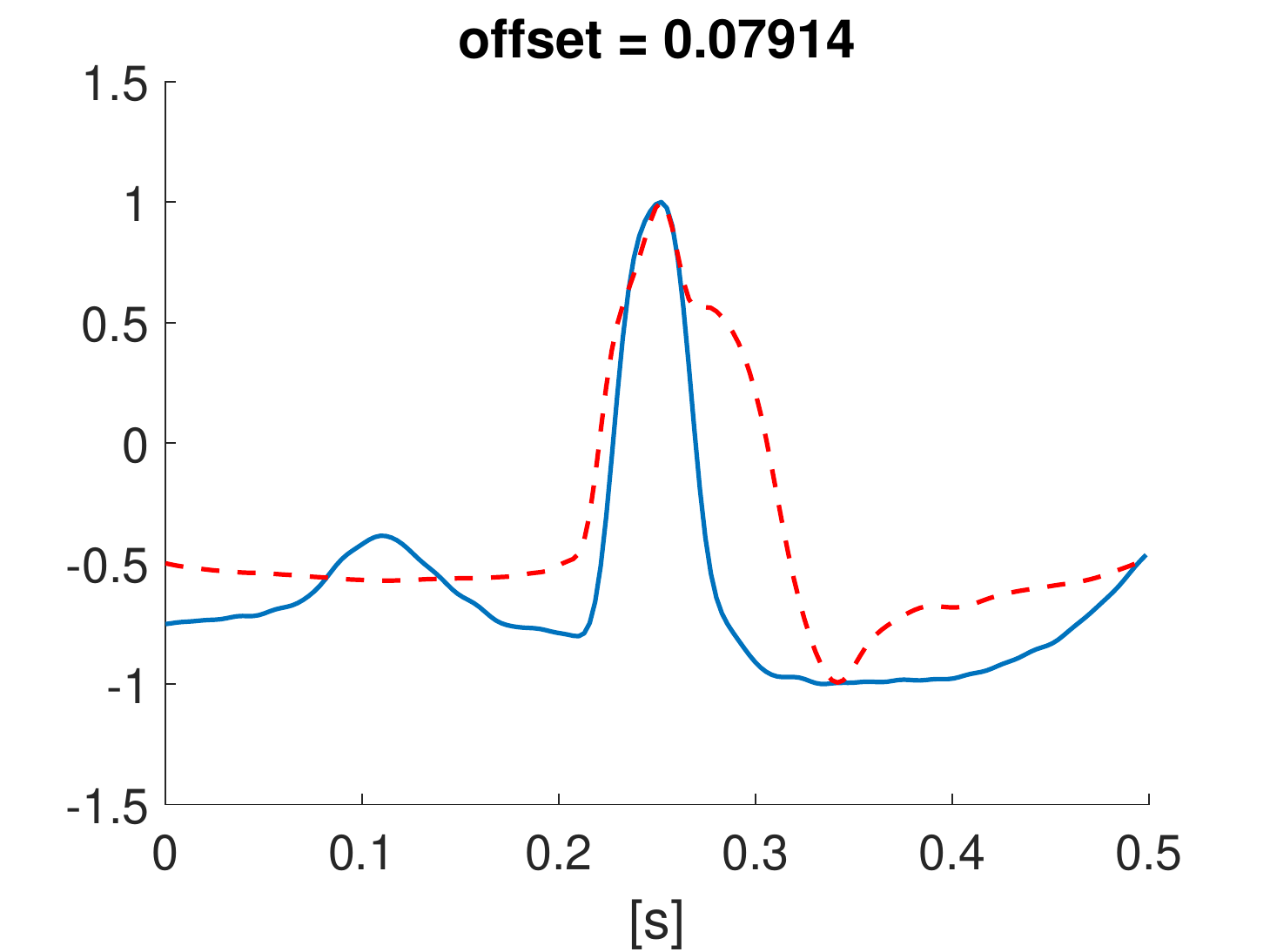}
\end{subfigure}\vspace{4pt}
\begin{subfigure}{.24\textwidth}
  \centering
  \includegraphics[width=1.2\linewidth]{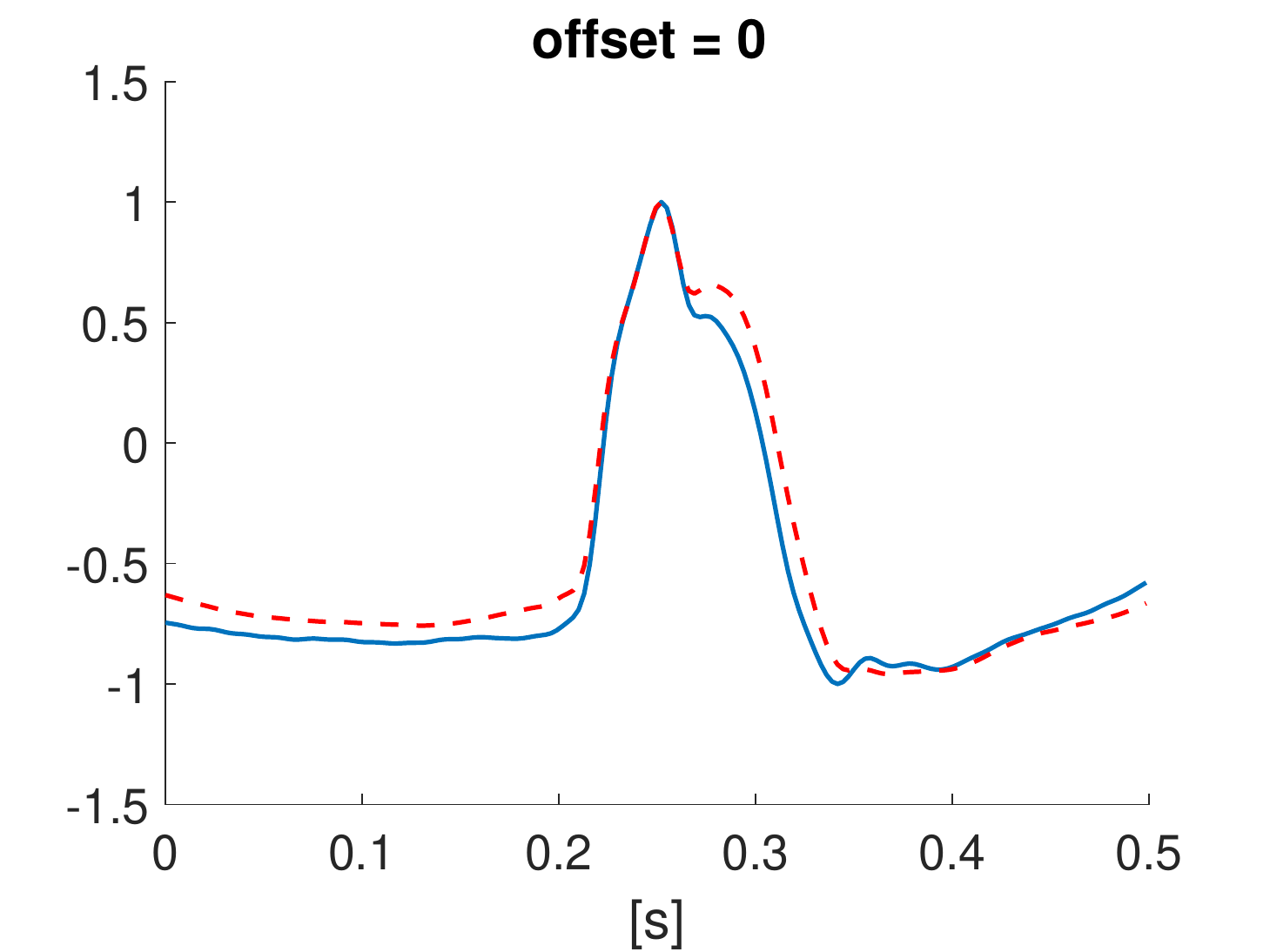}
\end{subfigure}\hfill
\begin{subfigure}{.24\textwidth}
  \centering
  \includegraphics[width=1.2\linewidth]{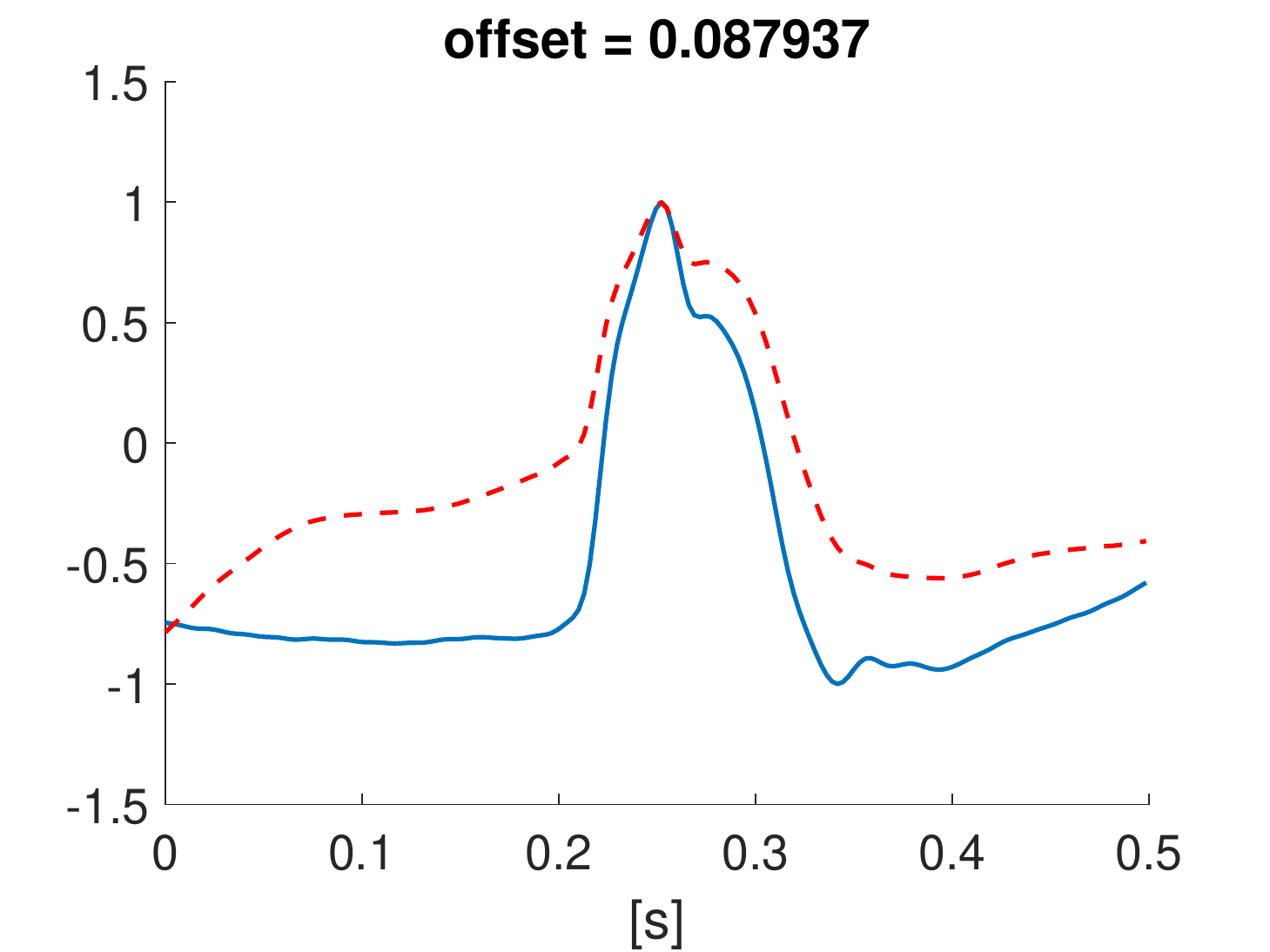}
\end{subfigure}\hfill
\begin{subfigure}{.24\textwidth}
  \centering
  \includegraphics[width=1.2\linewidth]{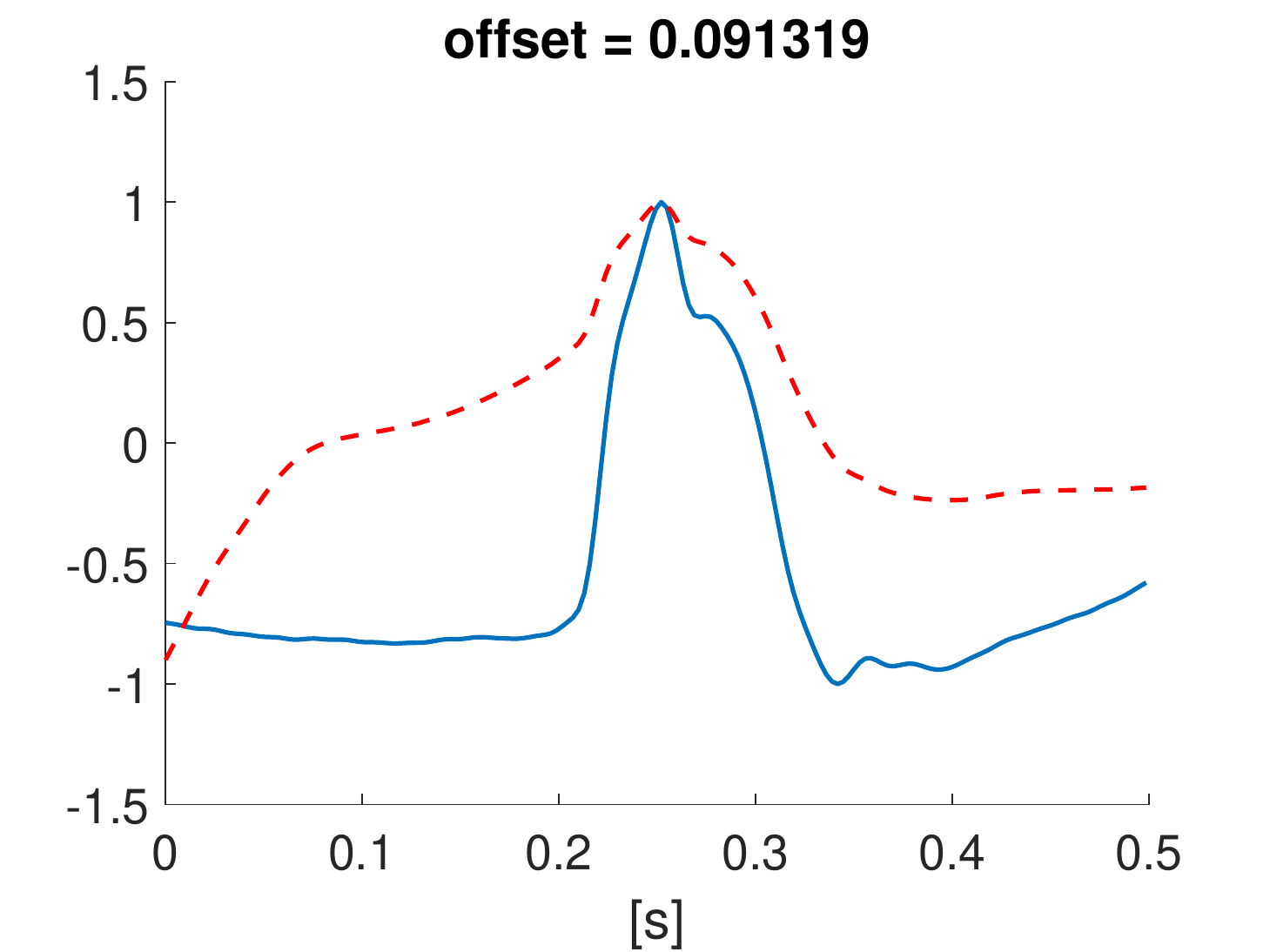}
\end{subfigure}\hfill
\begin{subfigure}{.24\textwidth}
  \centering
  \includegraphics[width=1.2\linewidth]{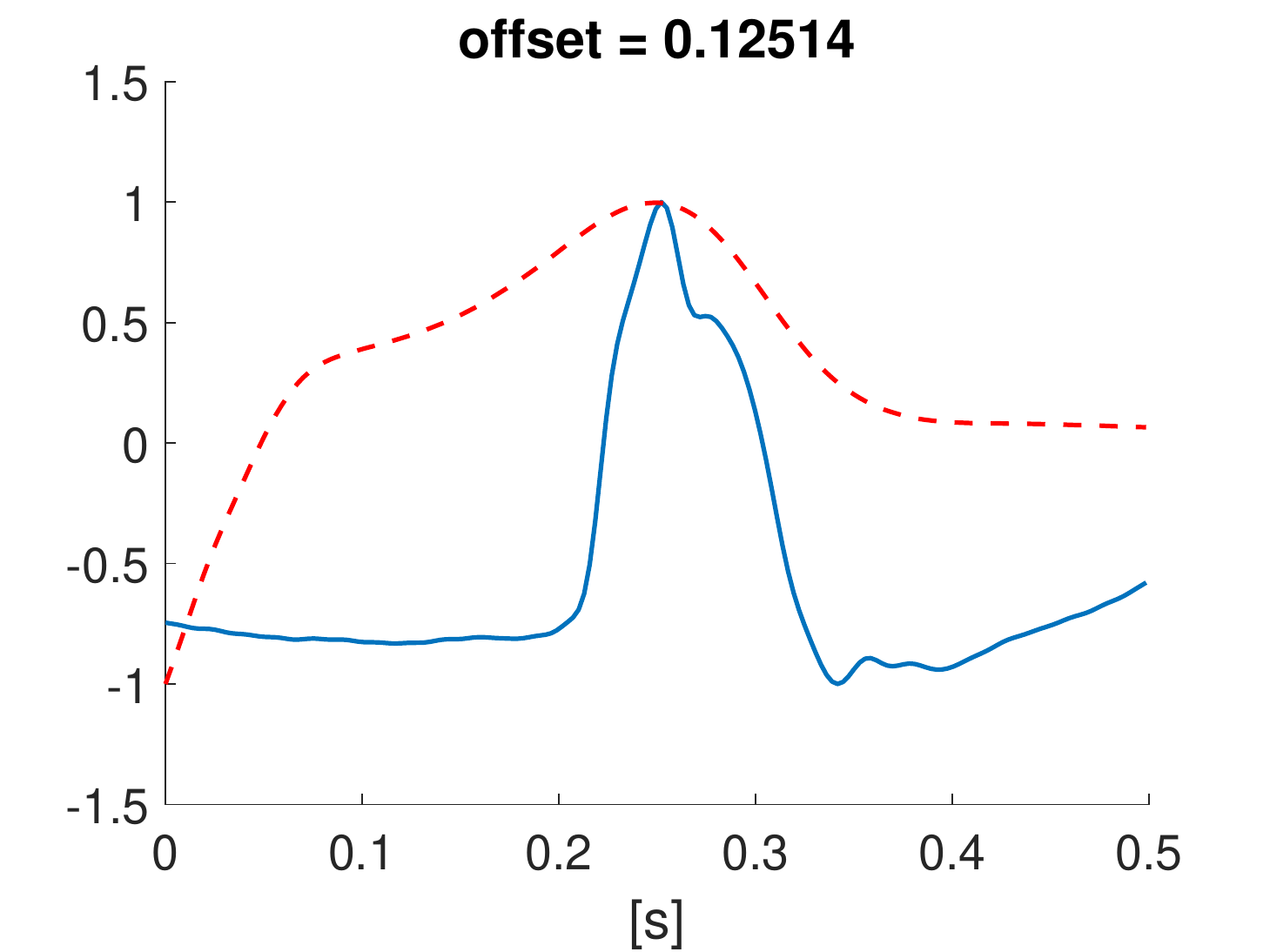}
\end{subfigure}\vspace{4pt}
\caption{Exploring the first three principal components of the latent feature space for the MIT-BIH arrhythmia database for normal and paced beats. The red line represents the newly generated datapoint compared to the original point depicted in blue. The top, middle and bottom row represent the variation along the first, second and third components respectively. Top and middle row start with a normal heartbeat pattern and the bottom row with a paced signal.}
\label{fig:explore_ECG}
\end{figure}

Fig. \ref{fig:explore_ECG} shows the result in input space of moving along the first principal components in the latent feature space. As original base point we take a normal beat signal, i.e. corresponding to a hidden unit on the bottom right of Fig. \ref{fig:ECG_feat_space_12}. The smooth transition between the beat patterns allows for interpretation of the first principal components. This allows a clinical expert to understand on what basis the paced beats are separated by the principal components and if this basis has a physiological meaning. In order to investigate the separated region of the latent space at the top of Fig. \ref{fig:ECG_feat_space_13} we start from a paced beat pattern and vary along the third principal component. This allows us to see which sort of heartbeat patterns are responsible for this specific distribution in the latent space.

\section{Novelty detection}
\label{sec:Outlier visualisation}

\begin{figure}[h]
    \centering
    \includegraphics[width=0.9\linewidth]{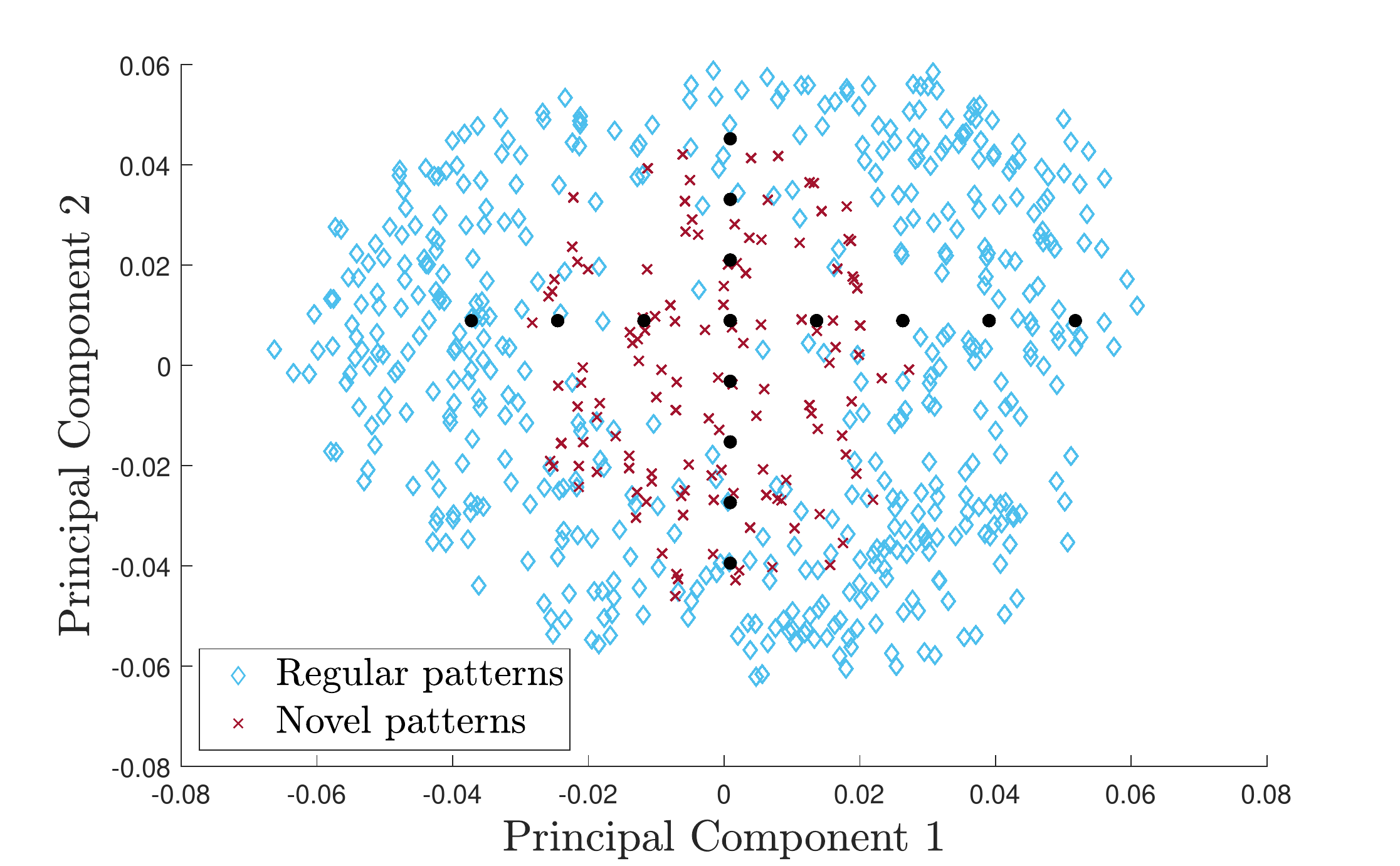}
    \caption{The latent space for 1000 zeros of the MNIST digits data set. The central cluster of points consists out of data points with a high novelty score, this corresponds to a low density region in the latent space. The black dots indicate the points in latent space which are sampled. }
    \label{fig:Novelty_feat_space}
\end{figure}

As a final illustration of latent space exploration using generative kernel PCA, we consider an application within the context of novelty detection. We use the reconstruction error in feature space as a measure of novelty~\cite{hoffmann2007kernel}, where Hoffmann shows the metric demonstrates a competitive performance on synthetic distributions and real-world data sets. The novelty score is calculated for all data points, where the 20\% of data points with the largest novelty score are considered novel. These points typically reside in low density regions of the latent space and are highlighted as interesting regions to explore using the tool. we consider 1000 instances of the digit zero from the MNIST dataset. After performing kernel PCA with the same parameters as in the previous section, we explore the latent space around the detected novel patterns. The data projected on the first two principal components is shown in Fig. \ref{fig:Novelty_feat_space}.

\begin{figure}[h]
\begin{subfigure}{.11\textwidth}
  \centering
  \includegraphics[width=.8\linewidth]{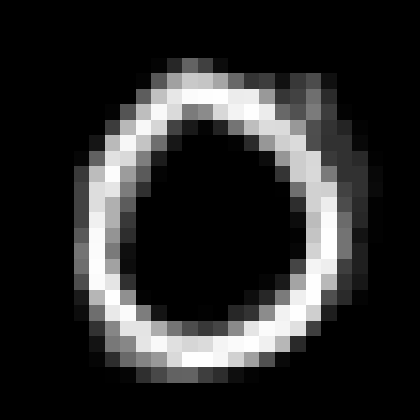}
\end{subfigure}
\begin{subfigure}{.11\textwidth}
  \centering
  \includegraphics[width=.8\linewidth]{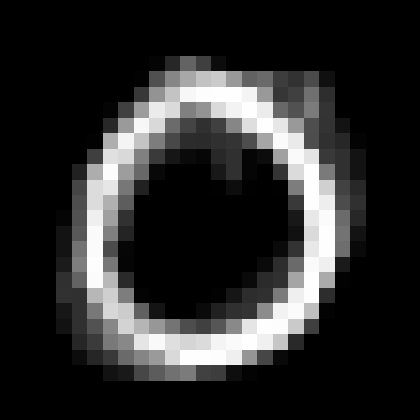}
\end{subfigure}
\begin{subfigure}{.11\textwidth}
  \centering
  \includegraphics[width=.8\linewidth]{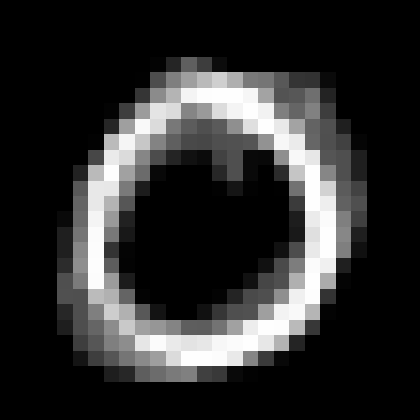}
\end{subfigure}
\begin{subfigure}{.11\textwidth}
  \centering
  \includegraphics[width=.8\linewidth]{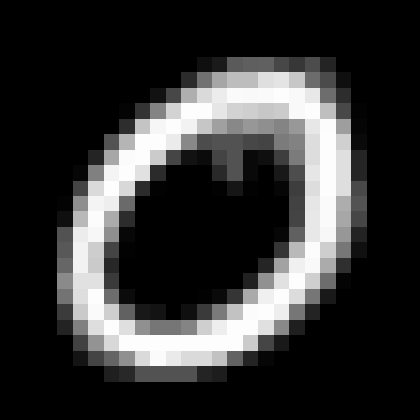}
\end{subfigure}
\begin{subfigure}{.11\textwidth}
  \centering
  \includegraphics[width=.8\linewidth]{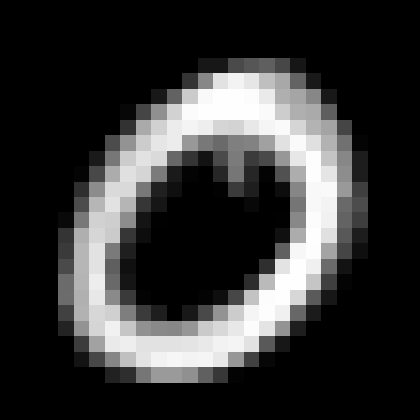}
\end{subfigure}
\begin{subfigure}{.11\textwidth}
  \centering
  \includegraphics[width=.8\linewidth]{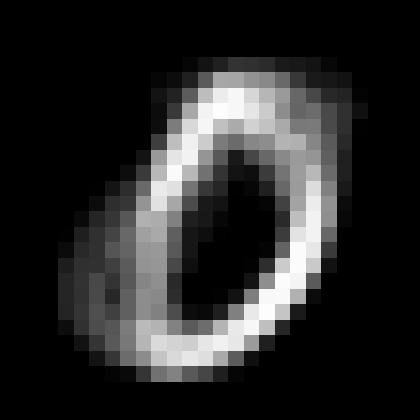}
\end{subfigure}
\begin{subfigure}{.11\textwidth}
  \centering
  \includegraphics[width=.8\linewidth]{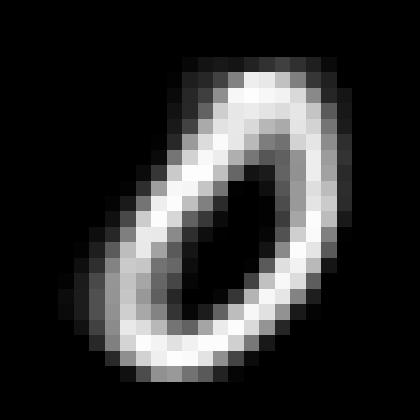}
\end{subfigure}
\begin{subfigure}{.11\textwidth}
  \centering
  \includegraphics[width=.8\linewidth]{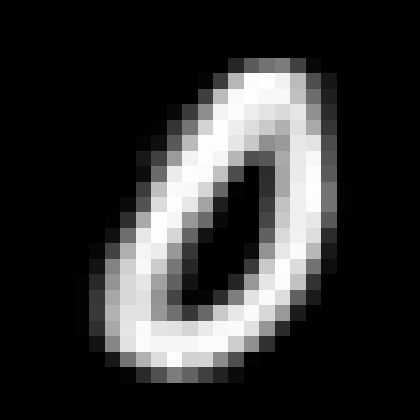}
\end{subfigure}\vspace{8pt}
\begin{subfigure}{.11\textwidth}
  \centering
  \includegraphics[width=.8\linewidth]{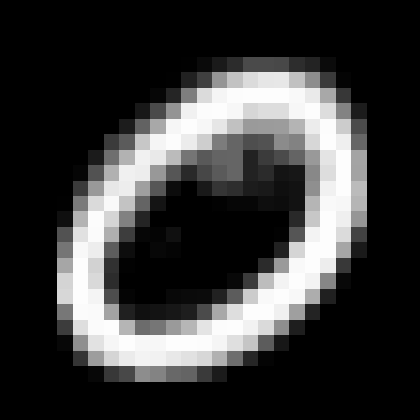}
\end{subfigure}\hfill
\begin{subfigure}{.11\textwidth}
  \centering
  \includegraphics[width=.8\linewidth]{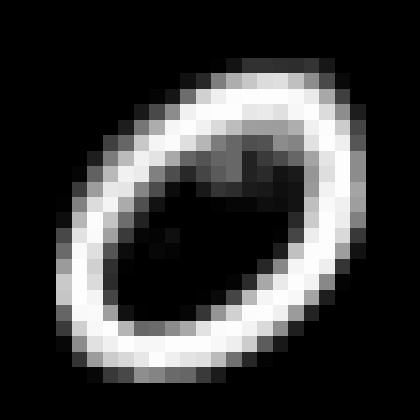}
\end{subfigure}\hfill
\begin{subfigure}{.11\textwidth}
  \centering
  \includegraphics[width=.8\linewidth]{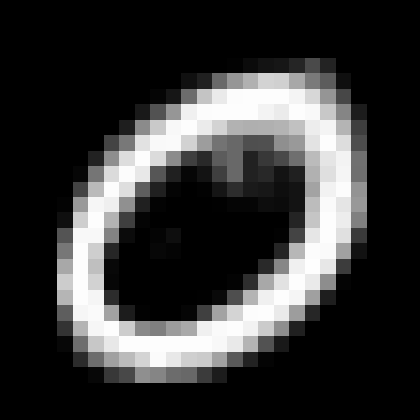}
\end{subfigure}\hfill
\begin{subfigure}{.11\textwidth}
  \centering
  \includegraphics[width=.8\linewidth]{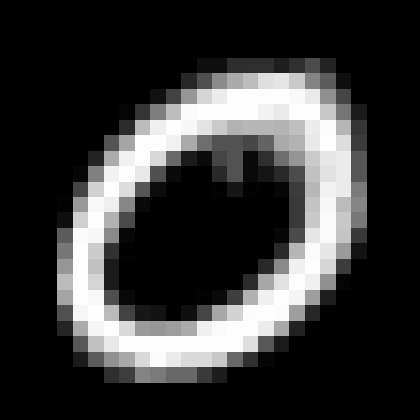}
\end{subfigure}\hfill
\begin{subfigure}{.11\textwidth}
  \centering
  \includegraphics[width=.8\linewidth]{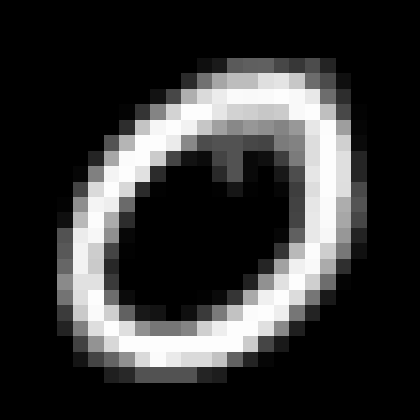}
\end{subfigure}\hfill
\begin{subfigure}{.11\textwidth}
  \centering
  \includegraphics[width=.8\linewidth]{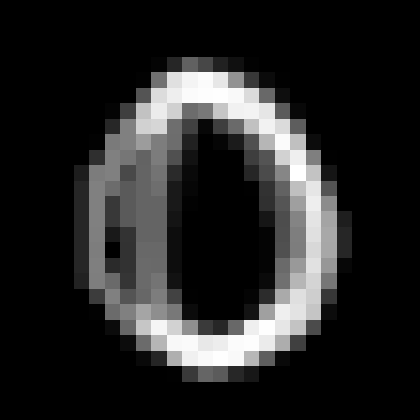}
\end{subfigure}\hfill
\begin{subfigure}{.11\textwidth}
  \centering
  \includegraphics[width=.8\linewidth]{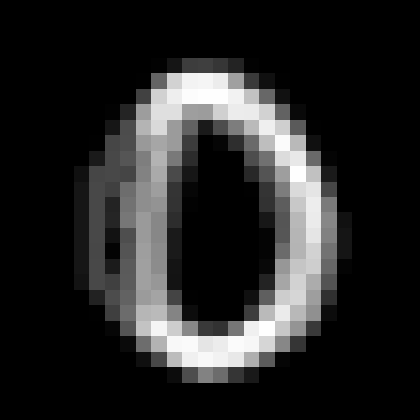}
\end{subfigure}\hfill
\begin{subfigure}{.11\textwidth}
  \centering
  \includegraphics[width=.8\linewidth]{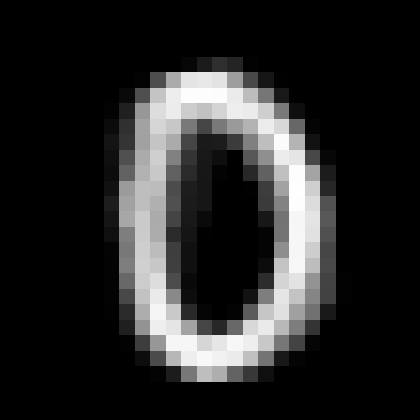}
\end{subfigure}\hfill
 \caption{Exploration of the latent space in Fig. \ref{fig:Novelty_feat_space}. The top row indicates the points generated from the horizontal black dots, while the bottom row correspond to the vertical positions.}
    \label{fig:Novelty_generated_digits}
\end{figure}

The generated images from the positions indicated by the black dots in Fig. \ref{fig:Novelty_feat_space} are shown in Fig. \ref{fig:Novelty_generated_digits}. The first row allows us to interpret the first principal component as moving from a thin round zero towards a more closed digit. The middle of the latent space is where the novel patterns are located which seems to indicate most zeros are either thin and wide or thick and narrow. A low amount of zeros in the data set are thick and wide or very thin and narrow. The bottom row of Fig. \ref{fig:Novelty_generated_digits} gives the interpretation for the second principal component as rotating the digit. The novel patterns seem to be clustered more together and as such have a less obvious orientation. 
Important to note is that we only look at the first 2 components for the interpretation, while in practice the novelty detection method takes all 20 components into consideration.

Above experiment shows that latent space exploration methods can give additional insights for novelty detection. Both the generating mechanism, as well as the novelty detection make use of the kernel PCA formulation. The two methods naturally complement each other: the novelty detection provides interesting regions in the latent space to explore, at the same time helps the generative mechanism in interpreting why certain points are considered as novel.

\section{Conclusion}
\label{sec:Conclusion}

The use of generative kernel PCA in exploring the latent space is demonstrated. Gradually moving along components in the feature space allows for the interpretation of components and consequently additional insight into the underlying latent space. This mechanism is demonstrated on the MNIST handwritten digits data set, the Yale Face Database B and the MIT-BIH Arrhythmia database. The last example showed generative kernel PCA to be a interesting method for obtaining an interpretable representation of the ECG beat embedding. As a final illustration, feature space exploration is used in the context of novelty detection~\cite{hofmann2008kernel}, where the latent space around novel patterns in data is explored. This to aid the interpretation of why certain points are considered as novel. Possible future directions would be the consideration of the geometry of the latent space. Not moving in straight lines, but curves through high density regions. Another direction would be to make use of different types of kernels as well as explicit feature maps for more flexibility in the latent feature space.  

\FloatBarrier

\subsection*{Acknowledgements}

EU: The research leading to these results has received funding from the European Research Council under the European Union’s Horizon 2020 research and innovation program / ERC Advanced Grant E-DUALITY (787960). This paper reflects only the authors’ views and the Union is not liable for any use that may
be made of the contained information. Research Council KUL: Optimization frameworks for deep kernel machines C14/18/068. Flemish Government: FWO: projects: GOA4917N (Deep Restricted Kernel Machines: Methods and Foundations), PhD/Postdoc grant. Flemish Government: This research received funding from the Flemish Government under the “Onderzoeksprogramma Artifici\"ele Intelligentie (AI) Vlaanderen” programme.

%
%
%
%

\bibliography{References}
\bibliographystyle{unsrt}

\end{document}